\documentclass[lettersize,journal]{IEEEtran}
\usepackage{amsmath,amsfonts}
\usepackage{algorithmic}
\usepackage{algorithm}
\usepackage{array}
\usepackage[caption=false,font=normalsize,labelfont=sf,textfont=sf]{subfig}
\usepackage{textcomp}
\usepackage{stfloats}
\usepackage{url}
\usepackage{verbatim}
\usepackage{graphicx}
\usepackage{cite}
\usepackage{booktabs}
\usepackage{multirow}
\usepackage{makecell}
\usepackage{pifont}
\newcommand{\cmark}{\ding{51}}
\newcommand{\xmark}{\ding{55}}
\usepackage[colorlinks=true,citecolor=blue,linkcolor=blue,urlcolor=blue]{hyperref}

\begin{document}

\title{\textbf{RainDancer}: RGB--Event Video Deraining with Rain-Oriented Spiking Dynamics}
%Decompose Before Interact: Component-Aligned RGB-Event Video Deraining}
%Event-Guided Video Deraining via Progressive Rain-Background Decoupling and Spiking Event Modeling}

 \author{Kui~Jiang,~\IEEEmembership{Member,~IEEE,}
         Runzhe~Li,
         Zhaocheng~Yu,
         Guanglu~Sun,~\IEEEmembership{Senior Member,~IEEE,}\\
         Junjun~Jiang,~\IEEEmembership{Senior Member,~IEEE,}
         Xianming~Liu,~\IEEEmembership{Senior Member,~IEEE}
        
\thanks{This research was financially supported by the National Natural Science Foundation of China (62501189), the Natural Science Foundation of Heilongjiang Province of China for Excellent Youth Project (YQ2024F006).} 
% the Open Research Fund from Guangdong Laboratory of Artificial Intelligence and Digital Economy (SZ) (GML-KF-24-09), and the Hubei Provincial Key Research and Development Program under Grant (2024BAB039).}
\thanks{Kui~Jiang, Runzhe~Li, Zhaocheng~Yu, Junjun~Jiang and Xianming~Liu are with the School of Computer Science and Technology, Harbin Institute of Technology, Harbin, China (e-mail: jiangkui@hit.edu.cn).}
\thanks{Guanglu~Sun is with the School of Computer Science and Technology, Harbin University of Science and Technology, Harbin, China (Sunguanglu@hrbust.edu.cn).}
}

\markboth{IEEE Transactions on Image Processing}%
{Kui \MakeLowercase{\textit{et al.}}: \textbf{RainDancer}: RGB--Event Video Deraining with Rain-Oriented Spiking Dynamics}
% \author{Anonymous Authors}

% \markboth{IEEE Transactions on Image Processing}{Anonymous Authors: Event-Guided Video Deraining}

% \IEEEpubid{0000--0000/00\$00.00~\copyright~2026 IEEE}

\maketitle

\begin{abstract}
Video deraining aims to recover clean visual content from rainy videos and is important for reliable perception under adverse weather. Existing video deraining methods mainly rely on RGB sequences and exploit temporal redundancy to recover rain-free contents. However, RGB-only restoration remains ambiguous in dynamic rainy scenes, where rain streaks, fine background textures, object boundaries, camera motion, and occlusions may exhibit similar visual patterns. Event cameras provide complementary motion-sensitive observations with high temporal resolution, making them useful for capturing thin and transient rain trajectories. Nevertheless, event streams also contain sensor noise and background-triggered responses. Direct RGB--Event fusion may therefore introduce cross-modal interference if rain-related and background-related factors are not explicitly distinguished. To address this issue, we propose RainDancer, a progressive RGB--Event video deraining framework based on a decompose-before-interact paradigm.
%we propose a progressive RGB--Event video deraining framework following a \emph{decompose-before-interact} principle. 
The key idea is to disentangle rain and background components within each modality before cross-modal interaction. In the RGB branch, frame features are progressively decomposed into rain and background representations. In the event branch, a rain-oriented spiking neural network module is developed to capture sparse and bursty event dynamics associated with rain motion. Component-level fusion is then performed between semantically aligned representations, including background-to-background fusion for structure preservation and rain-to-rain fusion for rain suppression. We further introduce event-domain supervision to regularize sparse event reconstruction, structural consistency, and gradient orientation, thereby improving the reliability of event guidance. Experiments on synthetic and real RGB--Event video deraining datasets show that our method achieves superior quantitative performance, visual quality, and downstream perception robustness compared with existing methods. Code is available at \url{https://github.com/AE86-plus/RainDancer}.
\end{abstract}


\begin{figure}[!t]
\centering
\includegraphics[width=\columnwidth]{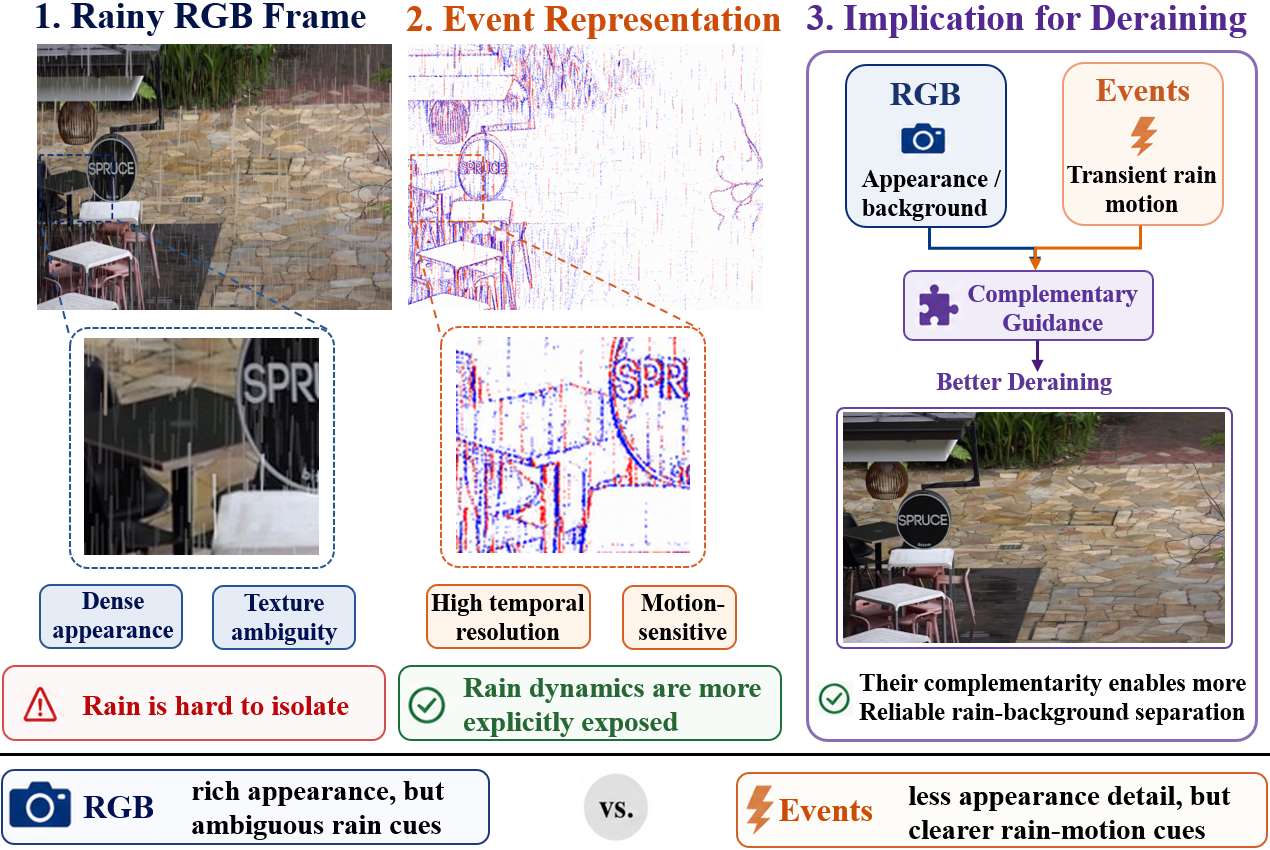}
\caption{Motivation of event guidance for rainy video restoration. In the RGB frame, rain streaks are easily entangled with background textures and wet-surface reflections, making them difficult to distinguish in the frame domain. In contrast, the event representation highlights sparse rain-induced dynamics, providing clearer motion-sensitive cues for rain-background separation.}
\label{fig:event_vs_rgb}
\end{figure}

\section{Introduction}
\IEEEPARstart{R}{ain} in outdoor videos introduces streak-like occlusions, contrast attenuation, background distortion, and temporal flickering~\cite{ref1}. These degradations not only reduce visual quality, but also impair downstream vision tasks such as object detection, tracking, semantic segmentation, and autonomous driving~\cite{yang2026cosos,dang2023efficient,xu2025unlocking}. Video deraining therefore plays an important role in reliable visual perception under adverse weather.

Existing deraining methods have been investigated in both single-image and video settings. Single-image deraining relies mainly on intra-frame spatial priors or learned representations to recover a clean image from a rainy observation~\cite{jiang2025ph,ref21,ref9}. However, without temporal information, rain streaks, fine textures, object boundaries, and motion-induced structures may exhibit similar local appearances, making rain-background separation ill-posed. Video deraining alleviates this ambiguity by exploiting temporal redundancy and inter-frame 
complementarity~\cite{ref4,ref7}. Early methods typically model rainy videos as the superposition of latent clean backgrounds and rain layers, and introduce handcrafted priors such as temporal coherence, chromatic consistency, low-rank background structure, directional sparsity, and stochastic rain statistics~\cite{ref6,ref32,ref33}. Although interpretable, these assumptions are often too restrictive for dynamic videos with heavy rain, camera motion, occlusions, and spatially non-uniform degradation.

Recent deep video deraining methods replace handcrafted modeling with data-driven spatio-temporal restoration. They improve rain removal through motion alignment, recurrent refinement, progressive reconstruction, flow regularization, self-aligned temporal aggregation, and long-range temporal modeling~\cite{ref7,ref8,ref36,ref41}. Some methods further consider complex rain effects such as rain accumulation, transmittance variation, and occlusion-related degradation~\cite{ref8,ref35}. Despite these advances, RGB-only video deraining remains challenging in dynamic rainy scenes. Rain streaks, background textures, object boundaries, camera motion, and occlusions may share similar spatial patterns or temporal variations. As a result, a model may mistakenly suppress true structures as rain or preserve rain artifacts as background details, especially under heavy rain or fast motion. This limitation suggests that temporal redundancy from RGB frames alone may be insufficient, and complementary sensing cues are desirable for more reliable rain-background discrimination.

Event cameras provide such a complementary modality. Unlike frame-based cameras, event cameras asynchronously record brightness changes at each pixel with high temporal resolution~\cite{ref10,ref11}. Each event is represented by its pixel location, timestamp, and polarity, and is triggered when the log-intensity variation exceeds a contrast threshold. This sensing mechanism produces sparse, low-latency, and motion-sensitive measurements, which have been widely used in motion-centric and restoration tasks, including optical flow estimation, camera tracking, event-to-frame reconstruction, motion deblurring, frame interpolation, super-resolution, and low-light enhancement~\cite{ref45,ref47,ref50,ref58}. These studies show that events can provide fine-grained temporal cues that are difficult to capture with conventional frame cameras.

The properties of event cameras are particularly relevant to video deraining. Rain streaks often induce rapid, localized, and non-uniform brightness changes, which may produce informative event responses. As illustrated in Fig.~\ref{fig:event_vs_rgb}, rain streaks that are visually entangled with background textures in RGB frames can become more distinguishable in the event domain. However, event streams are not clean rain indicators. They may also contain sensor noise, background-triggered events, and responses caused by camera or object motion. Therefore, the key challenge is not simply to inject events into RGB restoration, but to selectively exploit rain-related event cues while suppressing cross-modal interference.

\begin{figure}[!htb]
\centering
\includegraphics[width=\columnwidth]{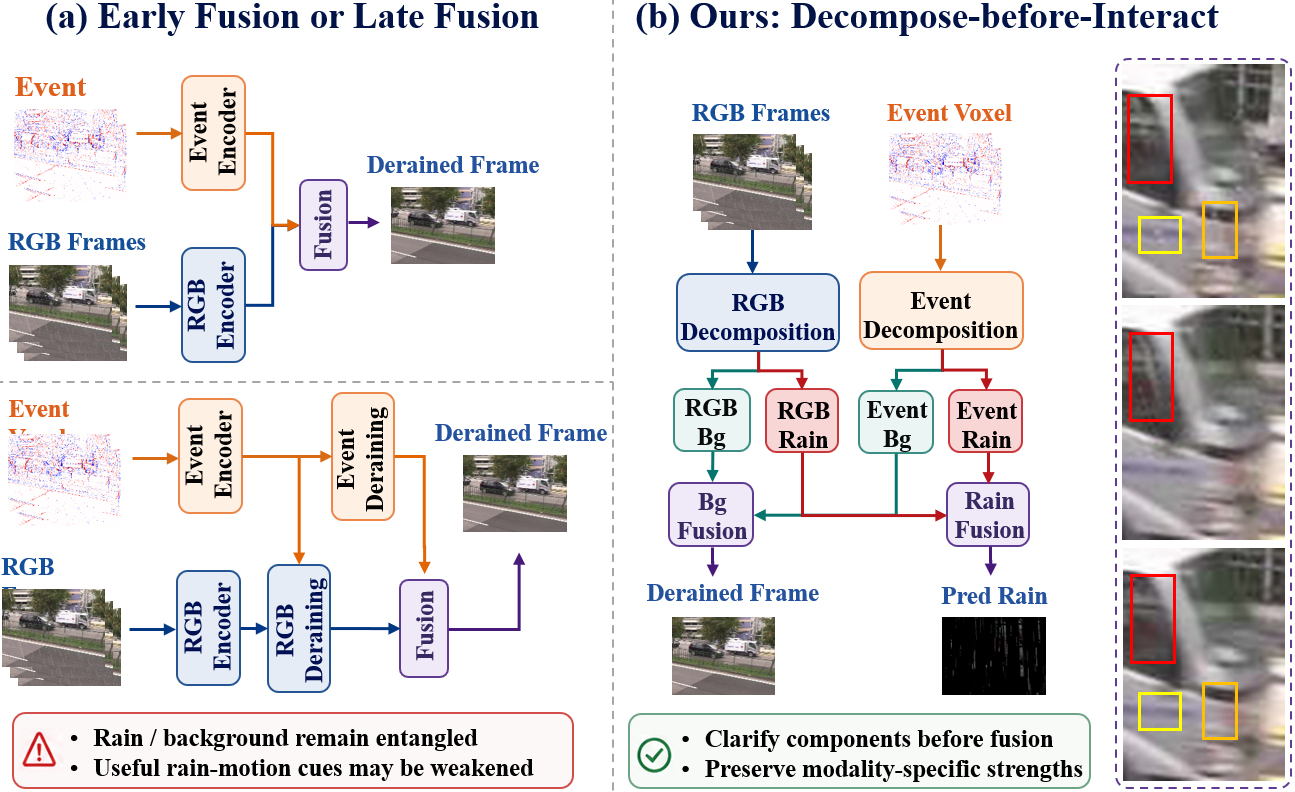}
\caption{Motivation comparison of RGB--Event interaction paradigms for video deraining. Early fusion directly mixes RGB appearance, rain-sensitive event responses, and background-triggered event noise, while late fusion may weaken transient event cues before they guide RGB restoration. In contrast, our framework first decomposes background and rain components within each modality, and then performs component-level interaction between semantically aligned components. The three results on the right correspond to MPEVNet, EGVD, and our RainDancer from top to bottom, respectively.}
\vspace{-2mm}
\label{fig:motivation_comparison}
\end{figure}

Recent event-guided video deraining methods have begun to explore RGB--Event collaboration~\cite{ref13,ref12}. MPEVNet introduces event features into a multi-stage RGB--Event fusion network and can be regarded as an early-fusion strategy~\cite{ref2}. EGVD adopts a late-fusion scheme, where modality-specific restoration is first performed and the restored representations are fused afterwards~\cite{ref12}. Other event-aware methods further exploit event representations for rain modeling or heterogeneous feature interaction~\cite{ref13,ref3}. These methods demonstrate the potential of event guidance, but their fusion mechanisms still operate mainly on mixed modality features. Early fusion may entangle RGB appearance, rain-sensitive event responses, and background-triggered event noise before rain and background factors are clarified. Late fusion, on the other hand, may weaken transient event cues before they can effectively guide RGB restoration. Consequently, existing fusion paradigms lack an explicit mechanism for aligning and interacting rain-related and background-related information across modalities. As shown in Fig.~\ref{fig:motivation_comparison}, residual rain artifacts and background structure loss remain visible in representative RGB--Event deraining results.

Spiking neural networks (SNNs) provide a natural computational model for event-based processing because they encode information through discrete spikes and temporal neuronal dynamics~\cite{ref15}. With surrogate-gradient training, residual structures, normalization, attention mechanisms, and spike-based Transformer architectures, SNNs have been extended from recognition to event-based perception and restoration tasks~\cite{ref62,ref65,ref69}. In adverse-weather imaging, SNNs have also been explored for event-driven deraining and rain-background decomposition~\cite{ref18,ref70}. These studies indicate that spiking dynamics are useful for modeling sparse and transient event responses. Nevertheless, most existing SNN-based deraining methods focus on event representation or direct RGB--Event fusion, without explicitly distinguishing the semantic roles of event responses. For event-guided video deraining, this distinction is important because rain-induced events and background-triggered events often coexist in the same event stream.

%we propose RainDancer, a progressive RGB--Event video deraining framework based on a decompose-before-interact paradigm.
To address these issues, we propose a \emph{decompose-before-interact} paradigm for event-guided video deraining. Instead of directly fusing heterogeneous RGB and event features, we first disentangle rain and background components within each modality, and then perform cross-modal interaction between semantically corresponding components. In this way, rain-related event responses can guide RGB rain removal, while background-related representations are fused separately to preserve scene structures. This design reduces the interference between rain cues and background cues while retaining the complementary advantages of both modalities.

Based on this principle, we propose RainDancer, a progressive RGB--Event video deraining framework. At each restoration stage, the RGB branch decomposes frame features into background and rain representations. For the event branch, we design a rain-oriented SNN module to model sparse and bursty event dynamics associated with rain motion, thereby separating rain-related and background-related event responses more explicitly. The two modalities are then coupled through component-level interactions, including background-to-background fusion and rain-to-rain fusion. In addition, we introduce event-domain supervision to regularize sparse event reconstruction, structural consistency, and gradient orientation, improving the reliability of event representations used for RGB restoration. As shown in Fig.~\ref{fig:motivation_comparison}, the proposed framework removes rain more thoroughly while preserving background structures more faithfully than representative RGB--Event fusion baselines.

The main contributions of this work are summarized as follows:
\begin{itemize}
\item We propose a progressive RGB--Event video deraining framework (RainDancer) based on a \emph{decompose-before-interact} principle. Different from conventional early or late fusion, RainDancer decomposes rain and background components before cross-modal interaction, thereby reducing premature feature entanglement and coarse representation aggregation.
\item We design a component-level RGB--Event interaction mechanism that separately performs background-to-background and rain-to-rain fusion. By aligning semantically corresponding components across modalities, the proposed mechanism strengthens event-guided rain removal while preserving background structures.
\item We develop a rain-oriented SNN event branch to capture sparse and bursty event dynamics induced by rain motion. Together with event-domain supervision on sparse reconstruction, structural consistency, and gradient orientation, the event branch provides more reliable event guidance for RGB video deraining.
\end{itemize}

\begin{figure*}[!tbp]
    \centering
    \includegraphics[width=\textwidth]{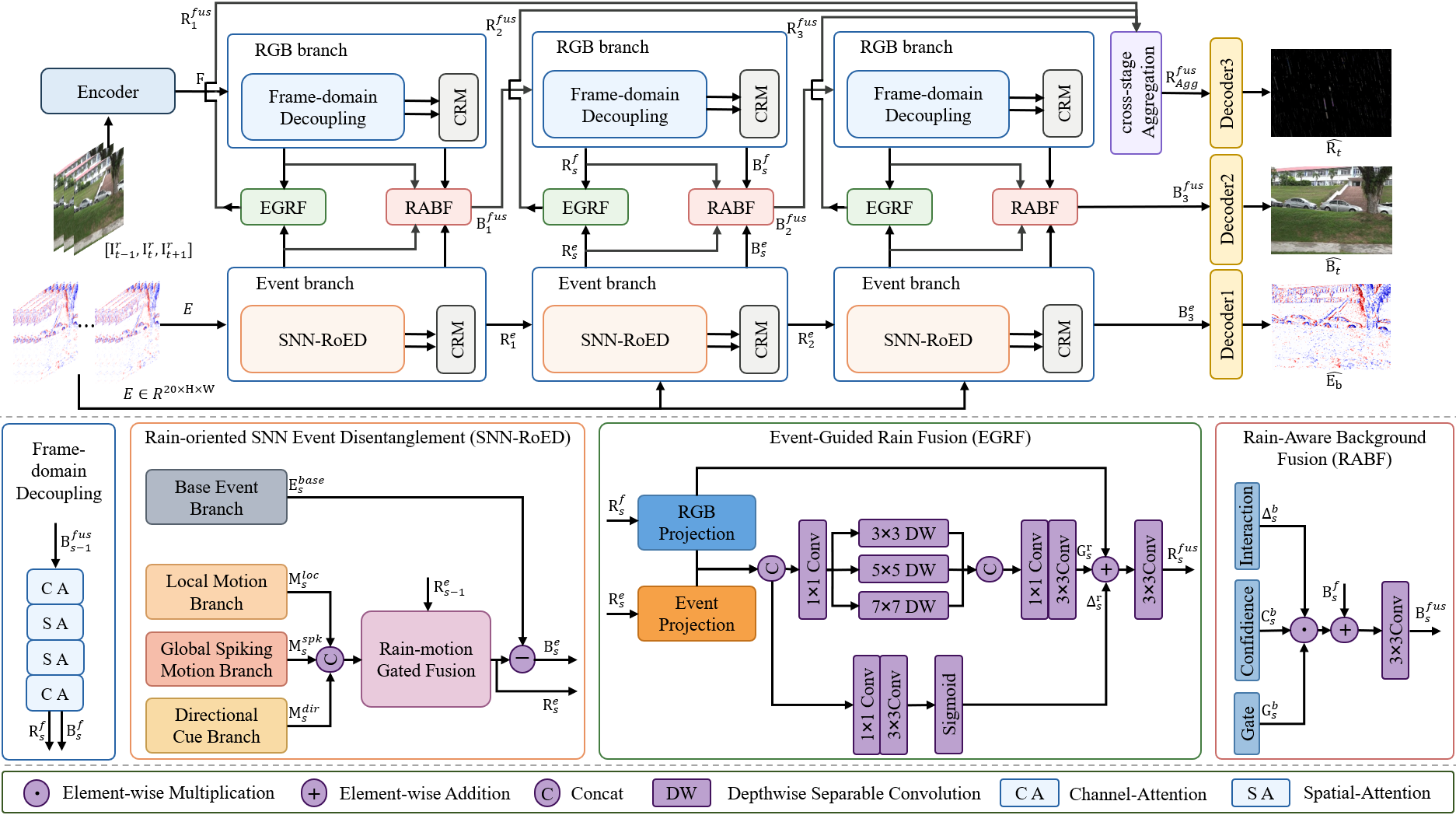}
    \caption{Overall architecture of the proposed RainDancer deraining framework. The network first extracts a spatio-temporal RGB feature with an Encoder block, and then performs three progressive stages. Each stage contains frame-domain decoupling, rain-oriented SNN event disentanglement (SNN-RoED), coupled representation modeling (CRM), event-guided rain fusion (EGRF), and rain-aware background fusion (RABF). The stage-wise rain features are collected by cross-stage aggregation, and the recovery module predicts the clean RGB background, clean event background, and RGB rain component.}\vspace{-4mm}
    \label{fig:framework}
\end{figure*}

\section{Proposed Model}

\subsection{Overview}

Fig.~\ref{fig:framework} illustrates the proposed progressive RGB--Event video deraining framework (RainDancer). Given a rainy RGB clip $\mathbf{X}=[\mathbf{I}_{t-1}^{r},\mathbf{I}_{t}^{r},\mathbf{I}_{t+1}^{r}]$ and the synchronized event tensor $\mathbf{E}\in\mathbb{R}^{20\times H\times W}$, the network predicts the restored RGB background $\hat{\mathbf{B}}_{t}$, the clean event background $\hat{\mathbf{E}}_{b}$, and the RGB rain component $\hat{\mathbf{R}}_{t}$. These outputs explicitly correspond to the background and rain factors considered in our rain-background decomposition formulation.

The framework follows a \emph{decompose-before-interact} principle. Instead of directly fusing RGB and event features, each modality first separates its representation into background and rain components. Cross-modal interaction is then performed between semantically aligned components, i.e., background-to-background fusion and rain-to-rain fusion. This design reduces the risk that rain-induced event responses or event noise are directly injected into RGB background restoration.
%The network follows a progressive decompose-before-interact pipeline. First, an Encoder+3D block extracts the initial spatio-temporal RGB feature $\mathbf{F}$ from neighboring frames. Then, three stages with the same topology progressively refine the representation. In the $s$-th stage, the RGB branch performs frame-domain decoupling to obtain RGB background and rain features, while the event branch uses the SNN-RoED module to disentangle event background and rain-motion features. Both modalities are then refined by coupled representation modules (CRMs). After intra-modal refinement, two cross-modal modules are applied: RERF fuses RGB/event rain features, and REBF uses reliable event background cues to update the RGB background feature under rain-aware constraints. The fused rain feature of each stage is stored for cross-stage aggregation, while the updated background feature is passed to the next stage.

The RGB clip is first encoded by an Encoder block to obtain an initial spatio-temporal feature $\mathbf{F}$. The following network contains three progressive stages with the same topology. At the $s$-th stage, the RGB branch decomposes the current RGB representation into background and rain features, while the event branch estimates event background and rain-motion features through a rain-oriented SNN event disentanglement module. The two pairs of features are then refined by coupled representation modules (CRMs). After intra-modal refinement, rain-level RGB--Event fusion and background-level RGB--Event fusion are performed by EGRF and RABF, respectively. The fused rain representation is stored for cross-stage aggregation, and the updated background representation is passed to the next stage.

%The stage-wise data flow can be written as
The stage-wise process is formulated as
\begin{equation}
\begin{aligned}
(\mathbf{B}^{f}_{s},\mathbf{R}^{f}_{s})
&=\mathrm{CRM}_{f}^{s}\!\left(D_{f}^{s}(\mathbf{B}^{\mathrm{fus}}_{s-1})\right), \\
(\mathbf{B}^{e}_{s},\mathbf{R}^{e}_{s})
&=\mathrm{CRM}_{e}^{s}\!\left(D_{e}^{s}(\mathbf{E},\mathbf{R}^{e}_{s-1})\right), \\
\mathbf{R}^{\mathrm{fus}}_{s}
&=\mathrm{EGRF}^{s}(\mathbf{R}^{f}_{s},\mathbf{R}^{e}_{s}), \\
\mathbf{B}^{\mathrm{fus}}_{s}
&=\mathrm{RABF}^{s}(\mathbf{B}^{f}_{s},\mathbf{B}^{e}_{s},
\mathbf{R}^{f}_{s},\mathbf{R}^{e}_{s}),
\end{aligned}
\end{equation}
where $D_{f}^{s}(\cdot)$ and $D_{e}^{s}(\cdot)$ denote frame-domain and event-domain decomposition, respectively. $\mathbf{B}^{\mathrm{fus}}_{s}$ is the updated background feature, $\mathbf{R}^{\mathrm{fus}}_{s}$ is the fused rain representation, and $\mathbf{R}^{e}_{s-1}$ is the event-rain prior propagated from the previous stage. For the first stage, $\mathbf{B}^{\mathrm{fus}}_{0}=\mathbf{F}$, and $\mathbf{R}^{e}_{0}$ is initialized as zero.

After three stages, the rain features $\{\mathbf{R}^{\mathrm{fus}}_{s}\}_{s=1}^{3}$ are aggregated into $\mathbf{R}^{\mathrm{fus}}_{Agg}$. A recovery module then takes $\mathbf{B}^{\mathrm{fus}}_{3}$, $\mathbf{R}^{\mathrm{fus}}_{Agg}$, and $\mathbf{B}^{e}_{3}$ as inputs and uses three lightweight decoders to predict $\hat{\mathbf{B}}_{t}$, $\hat{\mathbf{E}}_{b}$, and $\hat{\mathbf{R}}_{t}$, respectively.
%After the third stage, the rain features $\{\mathbf{R}^{\mathrm{fus}}_{s}\}_{s=1}^{3}$ are sent to the cross-stage aggregation module to form $\mathbf{R}^{\star}$. The recovery module then takes three inputs, including the last fused RGB background feature $\mathbf{B}^{\mathrm{fus}}_{3}$, the aggregated rain feature $\mathbf{R}^{\star}$, and the event background feature $\mathbf{B}^{e}_{3}$ from the third event branch. Its three decoders predict the clean RGB background $\hat{\mathbf{B}}_{t}$, the clean event background $\hat{\mathbf{E}}_{b}$, and the RGB rain component $\hat{\mathbf{R}}_{t}$, respectively.

\subsection{RGB Temporal Encoding and Frame-domain Decomposition}

% The RGB stream starts from the Encoder+3D block. A grouped 2D convolution first encodes neighboring frames with limited premature mixing, and a lightweight 3D convolution then aggregates short-range temporal information to produce $\mathbf{F}$. This design gives the following progressive stages a temporally aware RGB feature while keeping the front-end compact.

% At each stage, the frame-domain decoupling block separates the input RGB feature into background and rain components. As illustrated in Fig.~\ref{fig:framework}, the background branch extracts stable scene structures through consecutive channel attention (CA) and spatial attention (SA). CA adaptively emphasizes feature channels that are more related to background appearance and suppresses rain-sensitive responses, while SA further localizes spatially reliable background regions and weakens rain-contaminated areas. After obtaining the background feature, the rain branch is formed by a residual relation:
The RGB stream starts with a compact temporal encoder. A grouped 2D convolution first extracts frame-wise features from neighboring RGB frames while limiting premature cross-frame mixing. A lightweight 3D convolution then aggregates short-range temporal information and produces the initial feature $\mathbf{F}$. This front-end provides the progressive stages with temporally aware RGB representations.

At each stage, the frame-domain decomposition block separates the current RGB feature into background and rain components. The background branch estimates stable scene structures using channel and spatial attention. Channel attention emphasizes feature channels associated with persistent appearance, while spatial attention identifies regions where background cues are more reliable. Let $D_{f,b}^{s}(\cdot)$ denote the background estimation branch. The RGB decomposition is written as
\begin{equation}
\mathbf{B}^{f}_{s}=D_{f,b}^{s}(\mathbf{B}^{\mathrm{fus}}_{s-1}), \qquad
\mathbf{R}^{f}_{s}=\mathbf{B}^{\mathrm{fus}}_{s-1}-\mathbf{B}^{f}_{s}.
\end{equation}
The rain feature is obtained through a residual relation rather than an independent branch. This formulation is consistent with the additive rain-background decomposition assumption and constrains the two components to explain the input feature jointly.

Because rain and background may still be entangled after one-step decomposition, a CRM is used to refine the pair $\left(\mathbf{B}^{f}_{s},\mathbf{R}^{f}_{s}\right)$. The CRM exchanges information between the two streams, allowing residual rain responses in the background stream to be transferred to the rain stream and misclassified background details to be compensated back. This refinement improves the reliability of subsequent component-level RGB--Event interaction.
%This residual formulation follows the rain-background decomposition assumption and avoids learning two unrelated branches. Since rain streaks and background textures are still difficult to separate in a single pass, a CRM is applied after the initial decomposition. The CRM couples the two streams and exchanges complementary responses, so that rain activations remaining in the background stream can be transferred to the rain stream, while true background details mistakenly absorbed by the rain stream can be compensated back.

\subsection{Rain-oriented SNN Event Disentanglement}
Event streams provide motion-sensitive cues for rain, but they also contain background-triggered responses and sensor noise. Directly using such mixed event features for RGB restoration can introduce cross-modal interference. We therefore design a rain-oriented SNN event disentanglement module, termed SNN-RoED, to first enhance rain-related event dynamics and then estimate event background features by residual decomposition.

Given the event tensor $\mathbf{E}$, a base event branch extracts a compact event representation $\mathbf{E}^{\mathrm{base}}_{s}$. In parallel, three rain-motion branches are used to capture complementary event cues. The local motion branch uses 3D convolutions and temporal pooling to aggregate short-term burst responses caused by fast rain streaks. The global spiking motion branch employs 3D convolutions followed by an SNNformer adapted from SpikeVideoFormer~\cite{ref20}, which models long-range temporal dependencies in sparse event streams. The directional branch aggregates pre-event and post-event responses and applies 2D convolutions to enhance direction-aware rain traces.

The three motion cues are fused by a rain-motion gated unit. The event-rain estimate from the previous stage is used as a progressive prior:
\begin{equation}
\begin{aligned}
\mathbf{M}^{e}_{s}
&=\Phi_{\mathrm{gate}}^{s}
\left(
[\mathbf{M}^{\mathrm{loc}}_{s},
\mathbf{M}^{\mathrm{spk}}_{s},
\mathbf{M}^{\mathrm{dir}}_{s}],
\mathbf{R}^{e}_{s-1}
\right), \\
\mathbf{R}^{e}_{s}
&=\Phi_{r}^{s}(\mathbf{M}^{e}_{s}), \\
\mathbf{B}^{e}_{s}
&=\Phi_{b}^{s}(\mathbf{E}^{\mathrm{base}}_{s}-\mathbf{R}^{e}_{s}).
\end{aligned}
\end{equation}
Here, $\mathbf{M}^{\mathrm{loc}}_{s}$, $\mathbf{M}^{\mathrm{spk}}_{s}$, and $\mathbf{M}^{\mathrm{dir}}_{s}$ denote local, spiking, and directional rain-motion cues, respectively. The residual form encourages the event background to be estimated after removing rain-dominant responses from the base event representation. The resulting pair $\left(\mathbf{B}^{e}_{s},\mathbf{R}^{e}_{s}\right)$ is further refined by an event-domain CRM before cross-modal fusion.
%Here, $\mathbf{M}^{\mathrm{loc}}_{s}$, $\mathbf{M}^{\mathrm{spk}}_{s}$, and $\mathbf{M}^{\mathrm{dir}}_{s}$ denote local, spiking, and directional rain-motion cues, respectively. This design makes the event branch more sensitive to rain-related motion before background estimation. Similar to the RGB branch, a CRM further refines the event background/rain pair for cross-modal fusion.

\subsection{Rain-Aware Background Fusion}
The background fusion module, RABF, updates the RGB background feature using event background cues under rain-aware constraints. Its inputs are the RGB background feature $\mathbf{B}^{f}_{s}$, the event background feature $\mathbf{B}^{e}_{s}$, the RGB rain feature $\mathbf{R}^{f}_{s}$, and the event rain feature $\mathbf{R}^{e}_{s}$. The background components provide complementary structural information, while the rain components indicate regions where event responses are likely dominated by rain motion. Therefore, RABF uses rain features to regulate the reliability of event-guided background compensation.
%REBF performs rain-aware background-level RGB-Event interaction. It takes four decoupled features as input: the RGB background feature $\mathbf{B}^{f}_{s}$, the event background feature $\mathbf{B}^{e}_{s}$, the RGB rain feature $\mathbf{R}^{f}_{s}$, and the event rain feature $\mathbf{R}^{e}_{s}$. The two background features provide complementary appearance and motion-sensitive structural information, while the two rain features indicate where event responses may be dominated by rain motion rather than reliable background cues. Therefore, rain information is explicitly used to regulate background fusion, making REBF a rain-aware fusion module rather than a plain background fusion block.

%As shown in Fig.~\ref{fig:framework}, REBF contains an interaction branch, a gate branch, and a confidence branch. The interaction branch first concatenates the RGB background, event background, their element-wise product, and their absolute difference. The product term describes cross-modal similarity, while the absolute difference highlights modality-specific discrepancy. The concatenated feature is then processed by convolutions to produce the event-guided background residual $\Delta^{b}_{s}$. The gate branch receives the four original features 
RABF contains three branches. The interaction branch computes cross-modal relations from $[\mathbf{B}^{f}_{s},\mathbf{B}^{e}_{s},
\mathbf{B}^{f}_{s}\odot\mathbf{B}^{e}_{s},
|\mathbf{B}^{f}_{s}-\mathbf{B}^{e}_{s}|],$
where the product term measures feature consistency and the absolute difference highlights modality discrepancy. Convolutional layers then predict an event-guided background residual $\Delta^{b}_{s}$. The gate branch takes $[\mathbf{B}^{f}_{s},\mathbf{B}^{e}_{s},\mathbf{R}^{f}_{s},\mathbf{R}^{e}_{s}]$ as input and predicts a rain-aware gate $\mathbf{G}^{b}_{s}$. The confidence branch estimates an event confidence map $\mathbf{C}^{b}_{s}$ from $[\mathbf{B}^{e}_{s},\mathbf{R}^{e}_{s}]$ to suppress unreliable event background cues.
%, which adaptively controls whether event guidance should be introduced according to both background consistency and rain contamination. The confidence branch takes $\mathbf{B}^{e}_{s}$ and $\mathbf{R}^{e}_{s}$ as input and estimates an event confidence map $\mathbf{C}^{b}_{s}$, so that event background cues with strong rain interference can be suppressed.

The background feature is updated by a gated residual rule:
\begin{equation}
\mathbf{B}^{\mathrm{fus}}_{s}=
\Phi_{b,o}^{s}\!\left(
\mathbf{B}^{f}_{s}
+
\beta_{b}\,
\mathbf{G}^{b}_{s}\odot
\mathbf{C}^{b}_{s}\odot
\Delta^{b}_{s}
\right),
\end{equation}
where $\beta_b$ is a learnable residual scale. This conservative update injects event information only when it is both rain-aware and confidence-weighted, preventing noisy or rain-dominated event responses from corrupting the dense RGB background representation.
%where $\beta_b$ is a learnable residual scale. This formulation injects only gated and confident event-guided residuals into the RGB background stream. As a result, reliable event structures can compensate the RGB background, while rain-triggered or noisy event responses are prevented from overwhelming dense RGB appearance.

\subsection{Event-guided Rain Fusion and Cross-stage Aggregation}

% RERF performs rain-level interaction. RGB rain features describe appearance-level degradation, while event rain features provide sparse and motion-sensitive rain cues. Therefore, RERF does not replace RGB rain with event rain. Instead, it projects the RGB rain feature and event rain feature into an aligned space, concatenates them, and uses multi-scale depthwise branches with kernel sizes $3$, $5$, and $7$ to capture rain streaks with different spatial extents. The module predicts an event-guided rain compensation $\Delta^{r}_{s}$ and a rain fusion feature $\mathbf{G}^{r}_{s}$, which are combined with the projected RGB rain feature to generate the fused rain representation.

% The fused rain representation is computed as
The rain fusion module, EGRF, performs rain-to-rain RGB--Event interaction. RGB rain features encode appearance-level degradation, whereas event rain features provide sparse and motion-sensitive rain cues. EGRF aligns the two rain representations and predicts event-guided rain compensation without replacing the RGB rain stream.

Specifically, $\mathbf{R}^{f}_{s}$ and $\mathbf{R}^{e}_{s}$ are first projected to a common feature space and concatenated. Multi-scale depthwise convolution branches with kernel sizes $3$, $5$, and $7$ are then used to capture rain streaks with different spatial extents. The module predicts an event-guided compensation term $\Delta^{r}_{s}$ and a fused rain feature $\mathbf{G}^{r}_{s}$. The final stage-wise rain representation is
\begin{equation}
\mathbf{R}^{\mathrm{fus}}_{s}
=
\Phi_{r,o}^{s}\!\left(
\mathbf{G}^{r}_{s}
+
\Delta^{r}_{s}
+
\Phi_{r,p}^{s}(\mathbf{R}^{f}_{s})
\right),
\end{equation}
where $\Phi_{r,p}^{s}(\cdot)$ projects the RGB rain feature and $\Phi_{r,o}^{s}(\cdot)$ generates the fused rain representation. This formulation anchors the fused rain feature to the RGB degradation representation while using event cues to enhance transient rain dynamics.
%where $\Phi_{r,p}^{s}(\cdot)$ denotes the projection of the RGB rain feature. This formulation keeps the fused rain representation anchored to the RGB rain stream while using event-guided compensation and fusion features to enhance rain-related motion cues. The final convolution $\Phi_{r,o}^{s}(\cdot)$ produces the stage-wise fused rain feature.

The stage-wise rain features are finally collected by the cross-stage aggregation module, which uses channel and spatial attention to adaptively combine complementary rain responses from all three stages. In this way, coarse rain structures captured in early stages and fine residual streaks refined in later stages are integrated into a more complete rain representation for final recovery.
%The aggregation module uses channel/spatial attention to select useful rain responses across stages. Earlier stages tend to capture coarse and strong rain structures, while later stages focus on finer residual rain streaks.

\subsection{Optimization Objectives}

%The network is trained with an image restoration loss and an event-specific loss:
The model is optimized by an image-domain loss and an event-domain loss:
\begin{equation}
\mathcal{L}
=
\mathcal{L}_{\mathrm{img}}
+
\mathcal{L}_{\mathrm{event}}.
\end{equation}

%Following PCNet~\cite{ref71}, the restored clean background is optimized using Charbonnier loss, edge loss, and SSIM similarity. The rain component is also weakly supervised through a residual reconstruction constraint to regularize rain-background decomposition. Following the implementation, the image loss is
The image loss supervises both RGB background restoration and rain-background decomposition. Following PCNet~\cite{ref71}, we use Charbonnier loss, edge loss, and SSIM similarity for the restored background. An auxiliary residual reconstruction term is applied to the estimated rain component to regularize decomposition. The image-domain loss is defined as
\begin{equation}
\begin{aligned}
\mathcal{L}_{\mathrm{img}}
={}&0.3(\mathcal{L}_{\mathrm{char}}^{B}
+0.2\mathcal{L}_{\mathrm{char}}^{I})
+0.2\mathcal{L}_{\mathrm{edge}}^{B} \\
&-0.15(\mathcal{L}_{\mathrm{ssim}}^{B}
+0.2\mathcal{L}_{\mathrm{ssim}}^{I}),
\end{aligned}
\end{equation}
where the residual reconstruction terms are assigned smaller weights because they serve as decomposition regularizers rather than the primary restoration objective.
%Here, the auxiliary residual terms are assigned smaller weights because they regularize decomposition rather than serve as the primary restoration objective.

%The event loss is designed specifically for the SNN event branch. A naive pixel-wise loss over the 20-bin event tensor is easily dominated by inactive locations, since event representations are highly sparse. Moreover, event polarity may vary across local motion and illumination changes, while deraining mainly requires reliable activity, structure, and directional cues in the event domain. Therefore, both the predicted event background and the ground-truth clean event tensor are first converted to absolute values, denoted as $\mathbf{E}_{p}$ and $\mathbf{E}_{g}$. The event loss consists of three complementary terms:
The event-domain loss supervises the predicted clean event background $\hat{\mathbf{E}}_{b}$. Since event tensors are sparse, a naive pixel-wise loss is dominated by inactive positions. Moreover, for event-guided deraining, the reliability of activity distribution, structural boundaries, and gradient orientation is more important than exact polarity matching. We therefore convert both predicted and ground-truth clean event tensors to their absolute values, denoted by $\mathbf{E}_{p}$ and $\mathbf{E}_{g}$, and define
\begin{equation}
\mathcal{L}_{\mathrm{event}}
=
\lambda_{\mathrm{rec}}\mathcal{L}_{\mathrm{rec}}
+
\lambda_{\mathrm{grad}}\mathcal{L}_{\mathrm{grad}}
+
\lambda_{\mathrm{dir}}\mathcal{L}_{\mathrm{dir}}.
\end{equation}

%First, we adopt a sparsity-aware bin-level reconstruction loss. Let
First, a sparsity-aware reconstruction loss emphasizes active event regions. Let$\mathbf{A}=\sum_{c}\mathbf{E}_{g}^{c}$ be the event activity map over all bins. The activity-adaptive weight is
\begin{equation}
\mathbf{W}=1+\alpha\left(\frac{\mathbf{A}}{\mathrm{mean}(\mathbf{A})+\epsilon}\right)^{\gamma}.
\end{equation}
The reconstruction term is
\begin{equation}
\mathcal{L}_{\mathrm{rec}}
=
\frac{\sum \mathbf{W}\odot \rho(\mathbf{E}_{p}-\mathbf{E}_{g})}
{\sum \mathbf{W}+\epsilon},
\end{equation}
where $\rho(\cdot)$ is the Charbonnier penalty.
%where $\rho(\cdot)$ denotes the Charbonnier penalty. This term prevents the loss from being overwhelmed by the large number of empty event positions and encourages the branch to preserve sharp event responses.

%Second, we impose segment-level structural consistency. The 20 event bins are divided into pre- and post-event segments and summed within each segment, which reduces the instability caused by bin-wise sparsity. Sobel gradients are then computed on the two segment maps. The gradient weight $\mathbf{W}_{g}$ is derived from the gradient magnitude of the ground-truth clean event segment maps, i.e., 
Second, a segment-level gradient loss preserves event structures. The 20 event bins are divided into pre- and post-event segments and summed within each segment to reduce bin-wise sparsity. Sobel gradients are computed on the segment maps. The gradient weight is defined as
\begin{equation}
\mathbf{W}_{g}=1+\eta\|\nabla\mathbf{E}_{g}\|,
\end{equation}
and the structural consistency term is
\begin{equation}
\begin{aligned}
\mathcal{L}_{\mathrm{grad}}
=
\frac{
\sum \mathbf{W}_{g}\odot
\rho(\nabla_x\mathbf{E}_{p}-\nabla_x\mathbf{E}_{g})
}{
\sum \mathbf{W}_{g}+\epsilon
}\\
\quad+
\frac{
\sum \mathbf{W}_{g}\odot
\rho(\nabla_y\mathbf{E}_{p}-\nabla_y\mathbf{E}_{g})
}{
\sum \mathbf{W}_{g}+\epsilon
}.
\end{aligned}
\end{equation}
%This constraint preserves clean event boundaries and background structural details, which are important for stable event-guided RGB restoration.

\begin{table*}[!htbp]
\caption{Quantitative Comparison on Three Datasets in Terms of PSNR and SSIM.\textit{Modality}: RGB denotes frame-only input, Event denotes event-only input, and RGB+Event denotes joint frame-event input.
The best and second-best results among deraining methods are highlighted in \textbf{bold} and \underline{underline}, respectively. FLOPs are re-measured on $256\times256$ patches under a unified setting.}
\label{tab:main_comparison_256}
\centering
\setlength{\tabcolsep}{0pt}
\renewcommand{\arraystretch}{1.12}
\footnotesize

\newcommand{\twometrics}[2]{%
\begin{tabular}{@{}c@{\hspace{18pt}}c@{}}#1 & #2\end{tabular}%
}

\resizebox{\textwidth}{!}{%
\begin{tabular}{@{}
c
@{\hspace{24pt}}
c
@{\hspace{24pt}}
c
@{\hspace{24pt}}
c
@{\hspace{24pt}}
c
@{\hspace{24pt}}
c
@{\hspace{24pt}}
c
@{}}
\toprule
\multirow{2}{*}{Method} & \multirow{2}{*}{Modality}
& NTURain
& RainSynLight25
& RainSynComplex25
& \multirow{2}{*}{Params}
& \multirow{2}{*}{FLOPs} \\
\cmidrule(l{0pt}r{24pt}){3-5}
&
& \twometrics{PSNR$\uparrow$}{SSIM$\uparrow$}
& \twometrics{PSNR$\uparrow$}{SSIM$\uparrow$}
& \twometrics{PSNR$\uparrow$}{SSIM$\uparrow$}
& & \\
\midrule
Rainy (no processing) & RGB & \twometrics{30.41}{0.9108} & \twometrics{25.23}{0.9099} & \twometrics{16.85}{0.6791} & -- & -- \\
(CVPR 2018) SpacCNN~\cite{ref7}        & RGB & \twometrics{33.11}{0.9475} & \twometrics{32.78}{0.9239} & \twometrics{21.21}{0.5854} & -- & -- \\
(TIP 2019) FastDeRain~\cite{ref31}      & RGB & \twometrics{30.30}{0.9274} & \twometrics{29.42}{0.8683} & \twometrics{19.25}{0.5385} & -- & -- \\
(TPAMI 2020) J4RNet~\cite{ref8}        & RGB & \twometrics{32.14}{0.9480} & \twometrics{32.96}{0.9434} & \twometrics{24.13}{0.7163} & 0.41M & 53.10G \\
(TIP 2022) TMICS~\cite{ref73}           & RGB & \twometrics{37.38}{0.9704} & \twometrics{36.65}{0.9689} & \twometrics{29.49}{0.8933} & 8.21M & 997.06G \\
(TPAMI 2022) MFGAN~\cite{ref76}         & RGB & \twometrics{38.92}{0.9764} & \twometrics{36.99}{0.9760} & \twometrics{32.70}{0.9357} & 34.16M & 525.31G \\
(TPAMI 2023) ESTINet~\cite{ref74}       & RGB & \twometrics{37.48}{0.9700} & \twometrics{36.12}{0.9631} & \twometrics{28.48}{0.8242} & 29.90M & 2312.70G \\
(CVPR 2023) DRSformer~\cite{ref72}      & RGB & \twometrics{36.93}{0.9591} & \twometrics{36.84}{0.9739} & \twometrics{31.61}{0.9258} & 33.66M & 440.66G \\
(MM 2024) RainMamba~\cite{ref40}        & RGB & \twometrics{37.87}{0.9738} & \twometrics{36.74}{0.9741} & \twometrics{32.65}{0.9361} & 38.73M & 219.73G \\
(CVPR 2025) VDMamba~\cite{ref41}        & RGB & \twometrics{\underline{39.74}}{\textbf{0.9791}} & \twometrics{37.90}{\underline{0.9819}} & \twometrics{32.92}{\underline{0.9447}} & 12.70M & 75.89G \\
(ICLR 2026) DeLiVR~\cite{ref75}         & RGB & \twometrics{38.85}{0.9764} & \twometrics{\underline{38.39}}{0.9761} & \twometrics{\underline{33.14}}{0.9187} & 6.53M & 53.89G \\
\midrule
(TIP 2023) MPEVNet~\cite{ref2}  & RGB+Event & \twometrics{38.91}{0.9744} & \twometrics{37.18}{0.9790} & \twometrics{33.05}{0.9440} & 1.59M & 3535.30G \\
(ICCV 2023) UVD-Event~\cite{ref13} & RGB+Event & \twometrics{35.04}{0.9580} & \twometrics{32.34}{0.9523} & \twometrics{28.77}{0.8860} & 7.11M & 179.26G \\
(IJCV 2024) EHN~\cite{ref3}  & RGB+Event & \twometrics{35.44}{0.9652} & \twometrics{32.35}{0.9497} & \twometrics{28.04}{0.8629} & 0.45M & 25.05G \\
(TNNLS 2025) EGVD~\cite{ref12} & RGB+Event & \twometrics{38.80}{0.9761} & \twometrics{37.04}{0.9762} & \twometrics{32.79}{0.9397} & 3.72M & 163.35G \\
\midrule
RainDancer-S~(Ours) & RGB+Event & \twometrics{39.30}{0.9750} & \twometrics{36.95}{0.9750} & \twometrics{32.64}{0.9378} & 1.40M & 318.74G \\
RainDancer~(Ours) & RGB+Event & \twometrics{\textbf{40.40}}{\underline{0.9789}} & \twometrics{\textbf{38.59}}{\textbf{0.9826}} & \twometrics{\textbf{34.56}}{\textbf{0.9569}} & 10.17M & 865.84G \\
\bottomrule
\end{tabular}%
}

\vspace{1mm}
\begin{minipage}{\textwidth}
\footnotesize
\end{minipage}\vspace{-4mm}
\end{table*}
Third, a gradient-direction loss encourages coherent event structure:
\begin{equation}
\mathcal{L}_{\mathrm{dir}}
=
\mathrm{mean}\!\left(
\mathbf{W}_{g}\odot
\left(
1-
\frac{\nabla\mathbf{E}_{p}\cdot\nabla\mathbf{E}_{g}}
{\|\nabla\mathbf{E}_{p}\|\,\|\nabla\mathbf{E}_{g}\|+\epsilon}
\right)
\right).
\end{equation}
In our implementation, $\lambda_{\mathrm{rec}}=1.0$, $\lambda_{\mathrm{grad}}=0.5$, and $\lambda_{\mathrm{dir}}=0.1$. The event-domain supervision constrains the SNN event branch to produce clean and structurally reliable event representations, which improves the stability of component-level RGB--Event fusion.
%The direction-consistency term encourages the predicted clean event representation to maintain coherent gradient orientation with the ground-truth clean events instead of only matching event intensity. In our implementation, $\lambda_{\mathrm{rec}}=1.0$, $\lambda_{\mathrm{grad}}=0.5$, and $\lambda_{\mathrm{dir}}=0.1$. Overall, this structured event supervision makes the SNN branch more stable and more useful for cross-modal background estimation.

\section{Experiments}

\subsection{Experimental Details}

\subsubsection{Datasets}
We evaluate the proposed RainDancer method on three video deraining benchmarks, including NTURain~\cite{ref7}, RainSynComplex25~\cite{ref8}, and RainSynLight25~\cite{ref35}. NTURain provides paired synthetic rainy/clean videos for full-reference evaluation and real-world rainy videos for generalization analysis. RainSynComplex25 contains 33 test sequences with dense and complex rain patterns, while RainSynLight25 contains 27 test sequences with relatively mild rain degradation. Since these benchmarks are frame-based, we generate the corresponding event streams using the v2e simulator~\cite{ref77}, which enables paired RGB-Event evaluation under controlled settings. We further use RainVID\&SS~\cite{ref2} to examine whether deraining benefits downstream object detection and semantic segmentation.

\subsubsection{Training Details}
Our RainDancer is implemented in PyTorch and trained on an NVIDIA RTX 3080 GPU. Each training sample contains three consecutive rainy RGB frames and the corresponding event voxel representation generated by v2e. The center frame is used as the restoration target. During training, paired RGB/event samples are randomly cropped into $128\times128$ patches, and the temporal length is fixed to 3. We optimize the network using Adam with an initial learning rate of $1\times10^{-4}$, $\beta_1=0.5$, and $\beta_2=0.999$. The batch size is set to 4, and the learning rate is decayed by cosine annealing.
%Our model is implemented in PyTorch and trained on an NVIDIA RTX 3080 GPU. Each training sample consists of three consecutive rainy RGB frames and the corresponding event voxel representation generated by v2e, where the center frame is used as the restoration target. During training, paired RGB/event samples are randomly cropped into $128\times128$ patches, and the sequence length is fixed to 3. We optimize the network using Adam with an initial learning rate of $1\times10^{-4}$, $\beta_1=0.5$, and $\beta_2=0.999$, with a batch size of 4, and decay the learning rate by cosine annealing. The training objective combines RGB-domain restoration supervision and event-domain supervision. The RGB loss includes Charbonnier, edge, and SSIM-based terms, while the event loss constrains the predicted clean event representation through weighted reconstruction, structural gradient consistency, and gradient-direction consistency.

\subsubsection{Evaluation Metrics}
For full-reference image quality evaluation, we report PSNR~\cite{huynh2008scope} and SSIM~\cite{ref88} on restored target frames. To assess perceptual quality in real-world scenes where clean references are unavailable, we use four no-reference image quality metrics: NIQE~\cite{ref84}, BRISQUE~\cite{ref85}, ILNIQE~\cite{ref86}, and CLIPIQA~\cite{ref87}. In addition to low-level restoration metrics, we evaluate the influence of deraining on downstream perception. For object detection, we report mAP-based metrics~\cite{ref89}, and for semantic segmentation, we report mIoU~\cite{ref90}. These metrics provide a more comprehensive evaluation of whether the restored videos are beneficial for subsequent visual understanding.

The training objective consists of RGB-domain restoration supervision and event-domain supervision. The RGB loss includes Charbonnier, edge, and SSIM-based terms. The event loss regularizes the predicted clean event representation using weighted reconstruction, structural gradient consistency, and gradient-direction consistency.

% \subsubsection{Baselines}
% For paired synthetic datasets, we report PSNR and SSIM~\cite{ref88} on restored target frames. For real-world videos without clean references, we adopt four no-reference image quality metrics, including NIQE~\cite{ref84}, BRISQUE~\cite{ref85}, ILNIQE~\cite{ref86}, and CLIPIQA~\cite{ref87}. To evaluate the influence of deraining on high-level perception, we report mAP-based metrics~\cite{ref89} for object detection and mIoU~\cite{ref90} for semantic segmentation.

% We compare our method with representative deraining methods from different categories. FastDeRain~\cite{ref31} is included as a classical model-based video deraining method, and DRSformer~\cite{ref72} is used as a strong single-image deraining baseline. For RGB-based video deraining, we compare with DeLiVR~\cite{ref75}, ESTINet~\cite{ref74}, RainMamba~\cite{ref40}, MFGAN~\cite{ref76}, VDMamba~\cite{ref41}, TMICS~\cite{ref73}, J4RNet~\cite{ref8}, and SpacCNN~\cite{ref7}. For event-guided deraining, we include MPEVNet~\cite{ref2}, EGVD~\cite{ref12}, UVD-Event~\cite{ref13}, and EHN~\cite{ref3}. These baselines cover model-based, single-image, RGB video-based, and event-guided deraining settings.

\subsubsection{Baselines}
We compare our RainDancer method with representative deraining approaches from different categories. FastDeRain~\cite{ref31} is included as a classical model-based video deraining method, and DRSformer~\cite{ref72} is used as a strong single-image deraining baseline. For RGB-based video deraining, we compare with DeLiVR~\cite{ref75}, ESTINet~\cite{ref74}, RainMamba~\cite{ref40}, MFGAN~\cite{ref76}, VDMamba~\cite{ref41}, TMICS~\cite{ref73}, J4RNet~\cite{ref8}, and SpacCNN~\cite{ref7}. For event-guided deraining, we include MPEVNet~\cite{ref2}, EGVD~\cite{ref12}, UVD-Event~\cite{ref13}, and EHN~\cite{ref3}. These methods cover model-based, single-image, RGB video-based, and RGB-Event deraining settings.

\subsection{Experiments and Analysis}

\subsubsection{Comparison on Synthetic Datasets}

We first evaluate the proposed method on NTURain, RainSynLight25, and RainSynComplex25. As reported in Table~\ref{tab:main_comparison_256}, our RainDancer obtains the highest PSNR on all three datasets. Compared with the second-best results, the PSNR gains are 0.66 dB on NTURain, 0.20 dB on RainSynLight25, and 1.42 dB on RainSynComplex25. The improvement is more pronounced on RainSynComplex25, where rain patterns are denser and more spatially complex. This result is consistent with the motivation of our RainDancer method: when RGB rain streaks, background textures, and motion-induced variations become difficult to distinguish, component-level RGB-Event interaction provides more explicit cues for rain/background separation.

We also report a lightweight variant, RainDancer-S. RainDancer-S shares parameters across progressive stages and uses a reduced channel width. It achieves performance comparable to strong RGB-Event baselines and obtains better results than MPEVNet on NTURain, while substantially reducing computational cost. This suggests that the proposed decomposition and component interaction strategy is not limited to a high-capacity model.
% We also provide a lightweight variant, RainDancer-S. It shares parameters across progressive stages and reduces the channel width, since our full model follows a multi-stage progressive deraining design. RainDancer-S achieves comparable performance to MPEVNet and performs better on NTURain, while using fewer parameters and reducing FLOPs by about 91.0\%.

\begin{figure*}[!tbp]
\centering
\includegraphics[width=\textwidth]{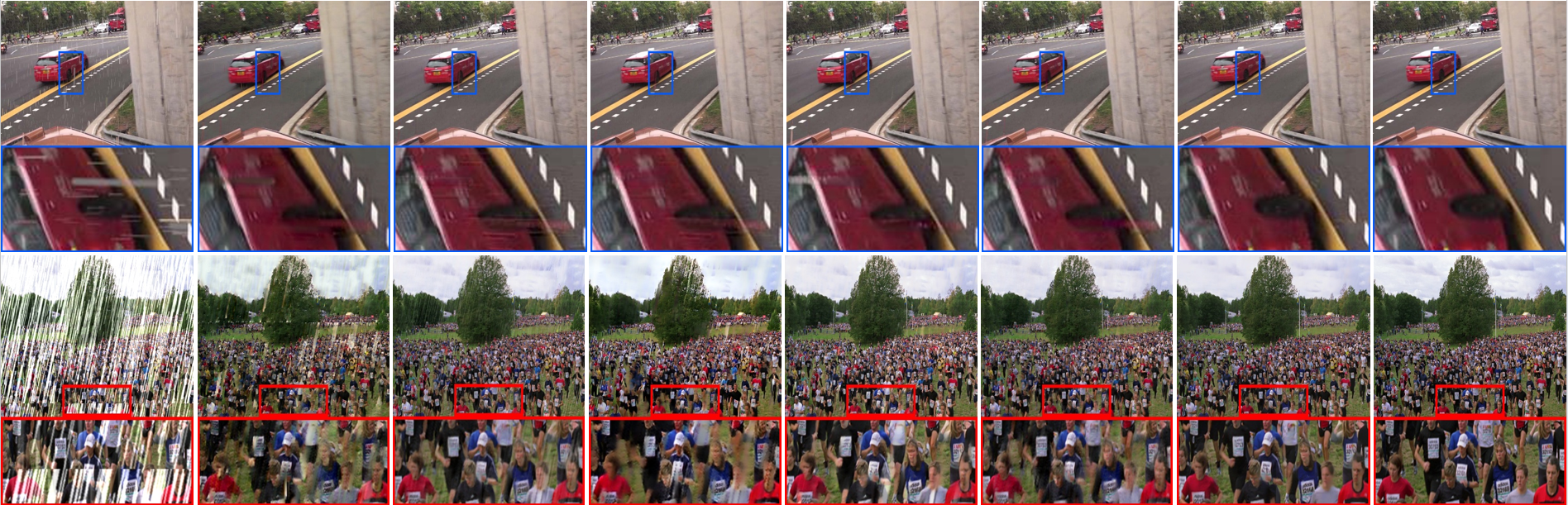}
\vspace{0.5mm}
\setlength{\tabcolsep}{0pt}
\newcommand{\figlabel}[2]{\makebox[0.125\textwidth][c]{\scriptsize\hspace{#1}#2}}
\noindent
\figlabel{-8pt}{Rainy}%
\figlabel{-8pt}{J4RNet~\cite{ref8}}%
\figlabel{-10pt}{ESTINet~\cite{ref74}}%
\figlabel{-8pt}{EGVD~\cite{ref12}}%
\figlabel{-8pt}{MPEVNet~\cite{ref2}}%
\figlabel{-8pt}{VDMamba~\cite{ref41}}%
\figlabel{-8pt}{RainDancer (Ours)}%
\figlabel{-8pt}{GT}
\caption{Qualitative comparison on synthetic datasets. From top to bottom, the examples are selected from NTURain and RainSynComplex25, respectively. Our results contain fewer artifacts.}\vspace{-4mm}
\label{fig:synthetic_comparison}
\end{figure*}

Fig.~\ref{fig:synthetic_comparison} shows qualitative comparisons on NTURain and RainSynComplex25. Several competing methods either leave residual streaks or introduce visible artifacts in regions with dense rain and background motion. In the NTURain example, some results contain color artifacts around rain-contaminated regions. In the RainSynComplex25 example, motion blur and incomplete rain removal are more evident. By contrast, our RainDancer method removes most rain streaks while maintaining sharper scene structures. These observations indicate that event cues are more effective when they are introduced after modality-specific rain/background decomposition, rather than being directly mixed with RGB appearance features.
% The qualitative results are shown in Fig.~\ref{fig:synthetic_comparison}. Our method removes rain while preserving high fidelity and avoiding excessive noise or artifacts. This advantage mainly comes from explicit rain-background decoupling and component-aligned RGB-Event interaction, which help the model use event rain cues without disturbing reliable RGB appearance. On NTURain, compared methods tend to introduce red artifacts, whereas our result better separates background motion from rain motion. On RainSynComplex25, our method produces less motion blur and preserves finer details than the compared methods.

\subsubsection{Comparison on Real-world Datasets}

We further evaluate real-world generalization on the real-world subset of NTURain. Since clean references are unavailable, Table~\ref{tab:nturain_real_nriqa} reports NIQE, BRISQUE, ILNIQE, and CLIPIQA. Our RainDancer method achieves the best BRISQUE, ILNIQE, and CLIPIQA scores, and ranks second on NIQE. DeLiVR obtains the best NIQE but is less competitive on BRISQUE and CLIPIQA. TMICS achieves a strong BRISQUE score, but its results tend to be smoother in detailed regions, as shown in Fig.~\ref{fig:realworld_comparison}. RainDancer-S performs slightly below the full model but remains competitive among both RGB-based and RGB-Event methods, indicating that the proposed design retains reasonable real-world robustness under a smaller model capacity.

\begin{table}[t]
\caption{No-reference IQA Comparison on the NTURain Real-World Dataset. Lower values indicate better perceptual quality for NIQE, BRISQUE, and ILNIQE, while higher values indicate better perceptual quality for CLIPIQA. The best and second-best results among deraining methods are highlighted in \textbf{bold} and \underline{underline}, respectively.}
\label{tab:nturain_real_nriqa}
\centering
\setlength{\tabcolsep}{3pt}
\renewcommand{\arraystretch}{1.08}
\scriptsize
\resizebox{\columnwidth}{!}{%
\begin{tabular}{c c c c c c}
\toprule
Method & Modality & NIQE$\downarrow$ & BRISQUE$\downarrow$ & ILNIQE$\downarrow$ & CLIPIQA$\uparrow$ \\
\midrule
Rainy (no processing) & RGB & 3.4565 & 22.9917 & 22.5554 & 0.4174 \\
(TPAMI 2020) J4RNet~\cite{ref8} & RGB & 3.4403 & 22.3564 & 27.3269 & 0.1943 \\
(TIP 2022) TMICS~\cite{ref73} & RGB & 3.5480 & \underline{18.5277} & 22.4265 & 0.4001 \\
(TPAMI 2022) MFGAN~\cite{ref76} & RGB & 3.3976 & 18.7521 & \underline{21.6317} & 0.3757 \\
(TPAMI 2023) ESTINet~\cite{ref74} & RGB & 3.4241 & 21.6854 & 22.2738 & 0.4145 \\
(CVPR 2023) DRSformer~\cite{ref72} & RGB & 3.5905 & 24.1146 & 22.8038 & 0.3929 \\
(MM 2024) RainMamba~\cite{ref40} & RGB & 3.5721 & 19.7478 & 22.0774 & 0.3848 \\
(CVPR 2025) VDMamba~\cite{ref41} & RGB & 3.3787 & 22.5778 & 22.4516 & 0.4117 \\
(ICLR 2026) DeLiVR~\cite{ref75} & RGB & \textbf{3.3118} & 20.3276 & 21.9527 & 0.3401 \\
\midrule
(TIP 2023) MPEVNet~\cite{ref2} & RGB+Event & 3.7146 & 21.3849 & 22.3000 & 0.3920 \\
(ICCV 2023) UVD-Event~\cite{ref13} & RGB+Event & 4.0071 & 19.3642 & 22.1203 & 0.3586 \\
(IJCV 2024) EHN~\cite{ref3} & RGB+Event & 4.8779 & 22.0881 & 32.9948 & 0.3716 \\
(TNNLS 2025) EGVD~\cite{ref12} & RGB+Event & 3.4212 & 22.7074 & 22.4844 & \underline{0.4310} \\
\midrule
RainDancer-S~(Ours) & RGB+Event & 3.3871 & 21.6012 & 21.9796 & 0.4198 \\
RainDancer~(Ours) & RGB+Event & \underline{3.3521} & \textbf{17.1424} & \textbf{21.5801} & \textbf{0.4328} \\
\bottomrule
\end{tabular}%
}
\end{table}

\begin{figure}[!t]
\centering
\includegraphics[width=\columnwidth]{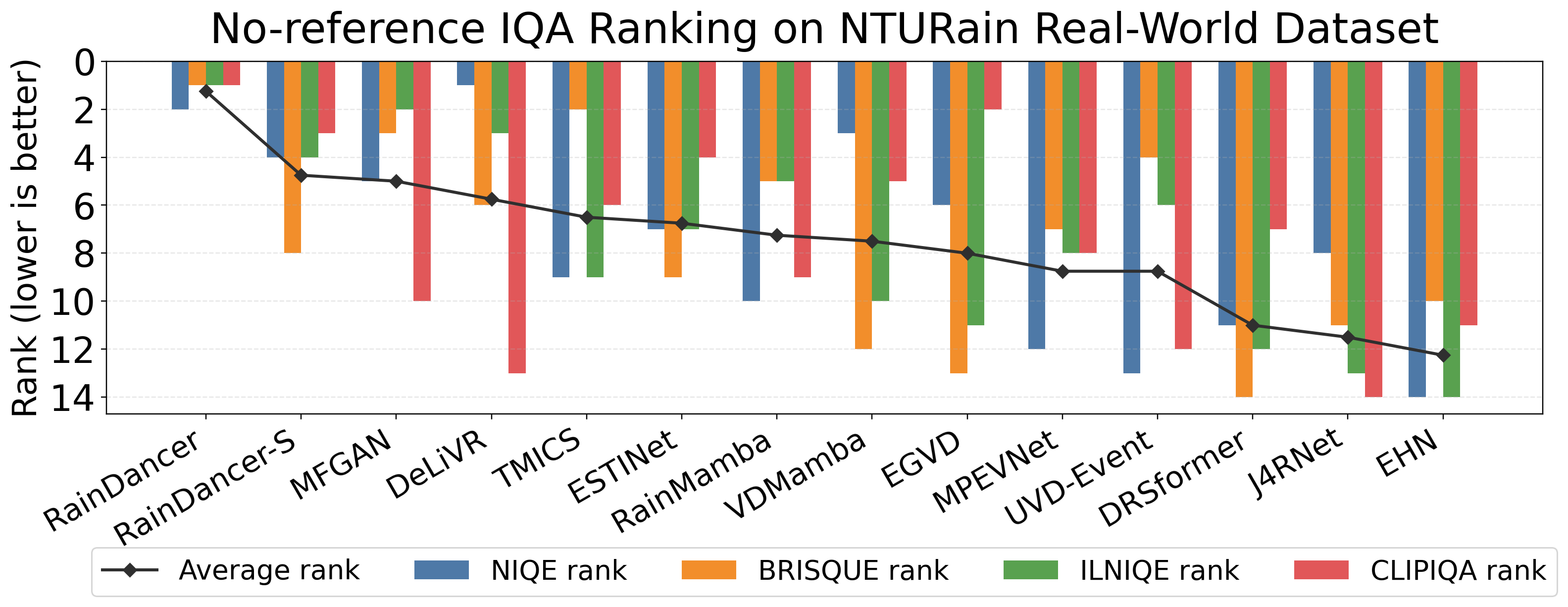}
\caption{Rank comparison of no-reference IQA metrics on the NTURain real-world dataset. Lower rank indicates better performance.}
\label{fig:realworld_nriqa_rankbar}
\end{figure}

Fig.~\ref{fig:realworld_nriqa_rankbar} summarizes the no-reference evaluation by ranking all methods on each metric and averaging the ranks. This rank-based view reduces the bias caused by relying on a single no-reference metric. Our RainDancer method obtains the best average rank and remains among the top methods across all four metrics, suggesting balanced perceptual behavior on real-world rainy videos. The lightweight RainDancer-S has a weaker BRISQUE rank than the full model but keeps a competitive average rank, which further supports the scalability of the proposed design.
% Fig.~\ref{fig:realworld_nriqa_rankbar} further summarizes the no-reference evaluation by ranking all methods on each metric and averaging the ranks. Compared with directly reading four raw metric values, the rank view makes the overall trend clearer. Our method achieves the best average rank and ranks highly on each individual metric, showing balanced real-world perceptual quality rather than improvement on only one score. RainDancer-S obtains a relatively weaker BRISQUE rank, but its average rank remains competitive, suggesting that the lightweight design preserves most of the robustness of the full model.

\begin{figure*}[!htbp]
\centering
\includegraphics[width=\textwidth]{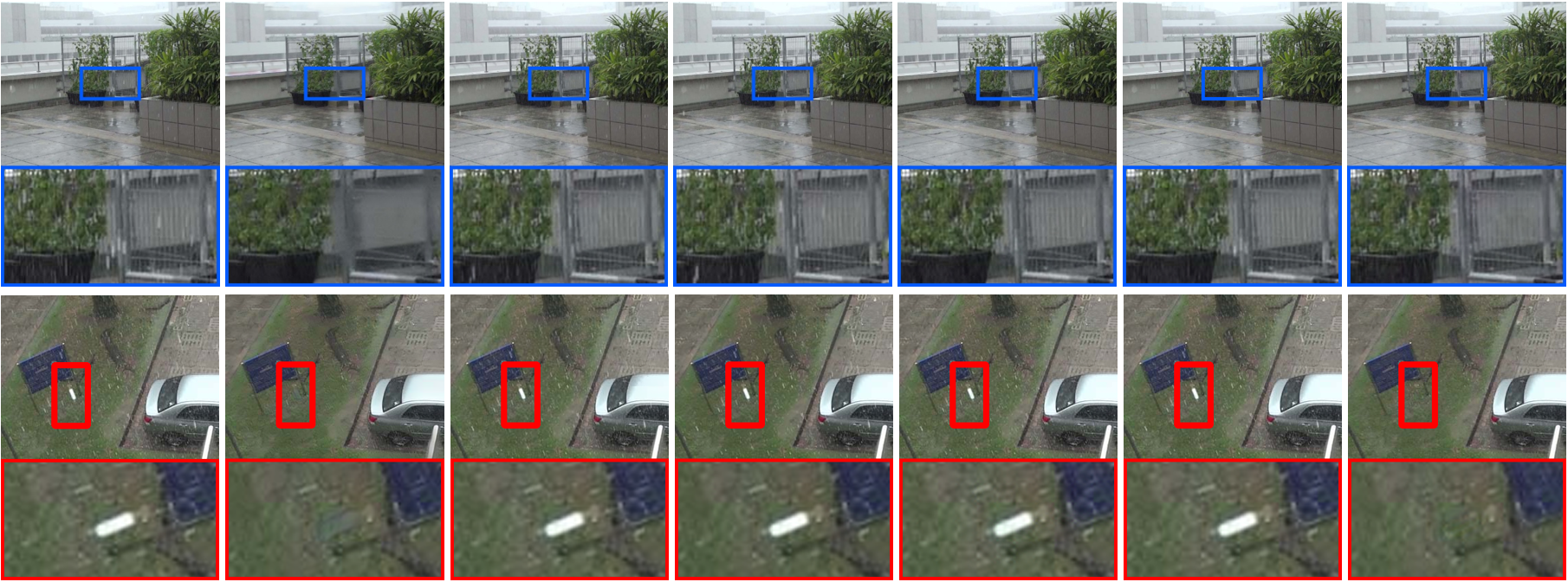}
\vspace{0.5mm}
\setlength{\tabcolsep}{0pt}
\newcommand{\realfiglabel}[2]{\makebox[0.1428\textwidth][c]{\scriptsize\hspace{#1}#2}}
\noindent
\realfiglabel{-8pt}{Rainy}%
\realfiglabel{-8pt}{TMICS~\cite{ref73}}%
\realfiglabel{-10pt}{RainMamba~\cite{ref40}}%
\realfiglabel{-8pt}{EGVD~\cite{ref12}}%
\realfiglabel{-8pt}{DeLiVR~\cite{ref75}}%
\realfiglabel{-8pt}{ESTINet~\cite{ref74}}%
\realfiglabel{-8pt}{RainDancer (Ours)}
\caption{Qualitative comparison on the NTURain real-world dataset. Our RainDancer produces clearer restored frames with fewer residual streaks.} \vspace{-4mm}
\label{fig:realworld_comparison}
\end{figure*}

\begin{table}[!t]
\centering
\caption{Ablation study on RGB-Event interaction paradigm.}
\label{tab:ablation_interaction}
\setlength{\tabcolsep}{7pt}
\renewcommand{\arraystretch}{1.1}
\footnotesize
\begin{tabular}{ccc}
\toprule
Variant & PSNR$\uparrow$ & SSIM$\uparrow$ \\
\midrule
RGB-only baseline & 30.28 & 0.9336 \\
Direct RGB-Event fusion & 35.64 & 0.9625 \\
Decompose-before-interact(Ours) & 40.40 & 0.9789 \\
\bottomrule
\end{tabular}
\end{table}
\begin{figure*}[!htbp]
\centering
\IfFileExists{ablation_interaction_visualization.png}{%
\includegraphics[width=\textwidth]{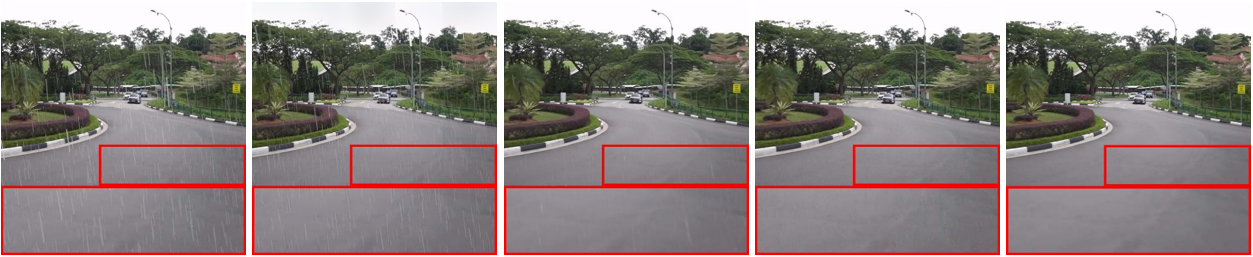}
}{%
\fbox{\parbox[c][0.16\textheight][c]{0.95\textwidth}{\centering Placeholder for interaction-paradigm visualization.}}
}
\vspace{0.5mm}
\setlength{\tabcolsep}{0pt}
\newcommand{\interfiglabel}[2]{\makebox[0.2\textwidth][c]{\scriptsize\hspace{#1}#2}}
\noindent
\interfiglabel{-8pt}{(a)}%
\interfiglabel{-8pt}{(b)}%
\interfiglabel{-8pt}{(c)}%
\interfiglabel{-8pt}{(d)}%
\interfiglabel{-8pt}{(e)}
\caption{Visualization for the interaction-paradigm ablation. (a) Rainy input, (b) RGB-only baseline, (c) direct RGB-Event fusion, (d) our decompose-before-interact method, and (e) ground truth. Our method better removes rain interference and produces results that are closest to the ground truth.} \vspace{-4mm}
\label{fig:ablation_interaction_visualization}
\end{figure*}
% Qualitative comparisons are shown in Fig.~\ref{fig:realworld_comparison}. TMICS produces visually strong deraining results, but it also introduces noticeable blur in detailed regions. DeLiVR also removes a large portion of real rain, but some short and bright rain streaks remain because they are likely treated as background structures. In contrast, our method removes rain streaks effectively while preserving scene structures and natural textures. This is mainly because the proposed decompose-before-interact design separates rain-related responses from background structures before RGB-Event interaction, allowing event cues to guide rain removal without introducing excessive noise or over-smoothing.

Fig.~\ref{fig:realworld_comparison} presents visual results on real-world rainy videos. TMICS removes many rain streaks but also smooths local textures. DeLiVR reduces obvious rain degradation, while some short bright streaks remain. EGVD and MPEVNet benefit from event guidance, but residual artifacts are still visible in regions with background motion. Our RainDancer produces cleaner results with fewer remaining streaks and less texture smoothing. This behavior is consistent with the component-level interaction strategy: rain-related event responses are mainly used to support rain removal, whereas background-related representations are fused separately to preserve scene structures.

% \subsubsection{Model Complexity Comparison}

% Model complexity is reported in Table~\ref{tab:main_comparison_256} in terms of parameters and FLOPs. For fair comparison, FLOPs are measured on $256\times256$ input patches using the common convention that one multiply-accumulate operation is counted as two floating-point operations. Our full model contains 10.17M parameters, which is much smaller than several strong RGB baselines such as ESTINet, MFGAN, and RainMamba, while achieving the best restoration performance on all three synthetic datasets. Its computational cost is higher than some lightweight RGB-only models due to the progressive RGB-Event decomposition design, but remains lower than TMICS, ESTINet, and MPEVNet. The lightweight RainDancer-S further improves the efficiency-performance trade-off. By sharing parameters across progressive stages and reducing the channel width, RainDancer-S uses only 1.40M parameters and 318.74G FLOPs. Compared with MPEVNet, it has fewer parameters and reduces the computational cost by about 11.1 times, while maintaining comparable performance and even achieving better results on NTURain. These results show that the proposed design can be scaled down effectively without losing its main advantage.

\subsubsection{Model Complexity Comparison}

Table~\ref{tab:main_comparison_256} also reports model complexity in terms of parameters and FLOPs. FLOPs are measured on $256\times256$ input patches, following the convention that one multiply-accumulate operation is counted as two floating-point operations. Our RainDancer contains 10.17M parameters, which is smaller than several strong RGB baselines such as ESTINet, MFGAN, and RainMamba. Its computational cost is higher than some lightweight RGB-only models because each progressive stage performs RGB-Event decomposition and component interaction. Nevertheless, its FLOPs remain lower than TMICS, ESTINet, and MPEVNet.

RainDancer-S provides a more efficient configuration. It uses 1.40M parameters and 318.74G FLOPs by sharing parameters across progressive stages and reducing the channel width. Compared with MPEVNet, RainDancer-S requires fewer parameters and reduces FLOPs by about 11.1 times, while maintaining comparable restoration accuracy and achieving better performance on NTURain. These results indicate that the proposed framework can be adjusted to different efficiency requirements without removing its core decomposition-interaction mechanism.

\subsection{Ablation Study}
All ablation variants are trained on NTURain using the same data split, optimizer, training schedule, and loss settings unless otherwise specified. We report PSNR and SSIM for quantitative comparison and provide visualizations to analyze how each component affects rain-background separation.

\subsubsection{Fusion Strategy}

We first study the RGB-Event interaction paradigm. Table~\ref{tab:ablation_interaction} compares three variants: an RGB-only baseline, a direct RGB-Event fusion baseline, and the proposed decompose-before-interact design. Direct RGB-Event fusion improves PSNR from 30.28 dB to 35.64 dB, confirming that events provide useful motion-sensitive information for video deraining. However, its performance remains clearly below our full design. This gap indicates that event information is not sufficient by itself; the way it is introduced into RGB restoration is critical.

As shown in Fig.~\ref{fig:ablation_interaction_visualization}, the RGB-only baseline leaves visible rain residues, especially where rain streaks overlap with background textures. Direct fusion reduces part of the rain degradation but still suffers from residual streaks and local artifacts, which suggests that background-triggered events may interfere with RGB restoration when mixed without explicit separation. Our RainDancer produces cleaner restoration because rain/background components are first separated within each modality and then interacted between semantically corresponding components. This result directly supports the proposed decompose-before-interact paradigm.

\subsubsection{Event Branch Processing}

We next evaluate the design of the event branch. Table~\ref{tab:ablation_snn_branch} compares a CNN event branch, a generic SNNformer event branch, and the proposed rain-oriented SNN branch. The CNN branch obtains 36.58 dB, indicating that a standard convolutional encoder can extract useful event features but has limited ability to model sparse temporal dynamics. Replacing it with SNNformer improves PSNR to 38.95 dB, showing the benefit of spike-driven temporal modeling. The proposed rain-oriented SNN further improves PSNR to 40.40 dB and SSIM to 0.9789, suggesting that adapting the event branch to rain-induced sparse and bursty responses is more effective than using a generic event encoder.

Fig.~\ref{fig:ablation_snn_event_features} visualizes the event features produced by different branches. The CNN branch gives diffuse responses and is easily affected by background-triggered events. SNNformer captures stronger activations but still responds to some non-rain structures. In comparison, the proposed rain-oriented SNN produces responses that are more concentrated on rain-related event patterns. This provides more discriminative event guidance for subsequent component-level RGB-Event interaction.
% The feature heatmaps in Fig.~\ref{fig:ablation_snn_event_features} further explain this improvement. The CNN branch produces relatively diffuse responses and is easily affected by background structures. SNNformer captures stronger event activations, but its responses are still not sufficiently concentrated on rain-induced motion. In contrast, our rain-oriented SNN shows clearer attention to rain-related event features, making rain streaks more distinguishable from background-triggered events. This verifies that the proposed event branch provides a stronger and more discriminative rain response, which benefits the subsequent RGB-Event interaction.

\begin{table}[!t]
\centering
\caption{Ablation study on the event branch design.}
\label{tab:ablation_snn_branch}
\setlength{\tabcolsep}{7pt}
\renewcommand{\arraystretch}{1.1}
\footnotesize
\begin{tabular}{ccc}
\toprule
Variant & PSNR$\uparrow$ & SSIM$\uparrow$ \\
\midrule
CNN & 36.58 & 0.9637 \\
SNNformer & 38.95 & 0.9748 \\
Rain-oriented SNN (Ours) & 40.40 & 0.9789 \\
\bottomrule
\end{tabular}
\end{table}

\begin{figure}[!t]
\centering
\IfFileExists{ablation_snn_event_features.png}{%
\includegraphics[width=\columnwidth]{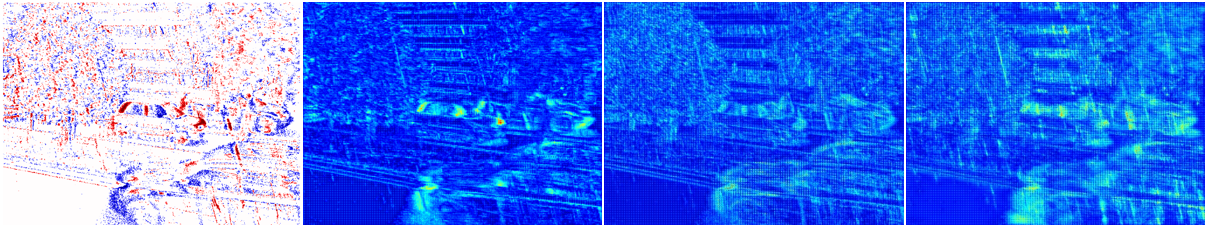}
}{%
\fbox{\parbox[c][0.16\textheight][c]{0.95\columnwidth}{\centering Placeholder for SNN event feature visualization.}}
}
\vspace{0.5mm}
\setlength{\tabcolsep}{0pt}
\newcommand{\snnfiglabel}[2]{\makebox[0.25\columnwidth][c]{\scriptsize\hspace{#1}#2}}
\noindent
\snnfiglabel{-8pt}{(a)}%
\snnfiglabel{-8pt}{(b)}%
\snnfiglabel{-8pt}{(c)}%
\snnfiglabel{-8pt}{(d)}
\caption{Feature visualization for the event branch ablation. (a) Event input, (b) CNN event branch, (c) SNNformer event branch, and (d) our rain-oriented SNN event branch.}\vspace{-2mm}
\label{fig:ablation_snn_event_features}
\end{figure}

\subsubsection{Loss Function}

We then analyze the event-domain supervision. Event tensors are sparse and contain many inactive pixels; therefore, a conventional pixel-wise loss may be dominated by zero-valued regions and provide insufficient constraints on meaningful event structures. Table~\ref{tab:ablation_event_loss} compares several supervision choices. The standard $\mathcal{L}_{1}+\mathcal{L}_{2}$ loss gives 39.12 dB. The sparsity-aware reconstruction loss $\mathcal{L}_{\mathrm{rec}}$ improves PSNR to 39.56 dB, indicating that event sparsity should be explicitly considered. The gradient consistency loss and direction consistency loss also improve the result, showing that local event structure and orientation are useful constraints for rain-related event modeling.

Combining the three terms further improves PSNR to 40.06 dB. The weighted formulation achieves the best result, 40.40 dB and 0.9789 SSIM, because it balances event reconstruction, structural consistency, and directional consistency. This confirms that event supervision contributes not only as an auxiliary regularizer but also as a way to improve the reliability of event representations used in RGB restoration.
% \subsubsection{Loss Function}

% We then investigate the design of event-domain supervision. Since event tensors are sparse and dominated by inactive pixels, directly applying a conventional image reconstruction loss cannot sufficiently constrain the predicted background events. Therefore, we compare several event losses in Table~\ref{tab:ablation_event_loss}, including the standard $\mathcal{L}_{1}+\mathcal{L}_{2}$ loss, each individual event loss term, their unweighted combination, and our weighted formulation. The reconstruction loss $\mathcal{L}_{\mathrm{rec}}$ already performs better than the standard pixel-wise loss, indicating that event sparsity should be explicitly considered. The gradient and direction losses further improve performance because they encourage structural and orientation consistency in the event domain. Combining all three terms gives stronger results, while the weighted formulation achieves the best PSNR and SSIM by balancing reconstruction fidelity, local event structure, and direction consistency.

\begin{table}[!t]
\centering
\caption{Ablation study on event-domain supervision.}
\label{tab:ablation_event_loss}
\setlength{\tabcolsep}{7pt}
\renewcommand{\arraystretch}{1.1}
\footnotesize
\begin{tabular}{ccc}
\toprule
Variant & PSNR$\uparrow$ & SSIM$\uparrow$ \\
\midrule
$\mathcal{L}_{1}+\mathcal{L}_{2}$ & 39.12 & 0.9745 \\
$\mathcal{L}_{\mathrm{rec}}$ & 39.56 & 0.9761 \\
$\mathcal{L}_{\mathrm{grad}}$ & 39.69 & 0.9766 \\
$\mathcal{L}_{\mathrm{dir}}$ & 39.67 & 0.9765 \\
$\mathcal{L}_{\mathrm{rec}}+\mathcal{L}_{\mathrm{grad}}+\mathcal{L}_{\mathrm{dir}}$ & 40.06 & 0.9781 \\
$\mathcal{L}_{\mathrm{rec}}+0.5\mathcal{L}_{\mathrm{grad}}+0.1\mathcal{L}_{\mathrm{dir}}$ (Ours) & 40.40 & 0.9789 \\
\bottomrule
\end{tabular}
\end{table}

\begin{figure}[!t]
\centering
\IfFileExists{ablation_event_loss_visualization.png}{%
\includegraphics[width=\columnwidth]{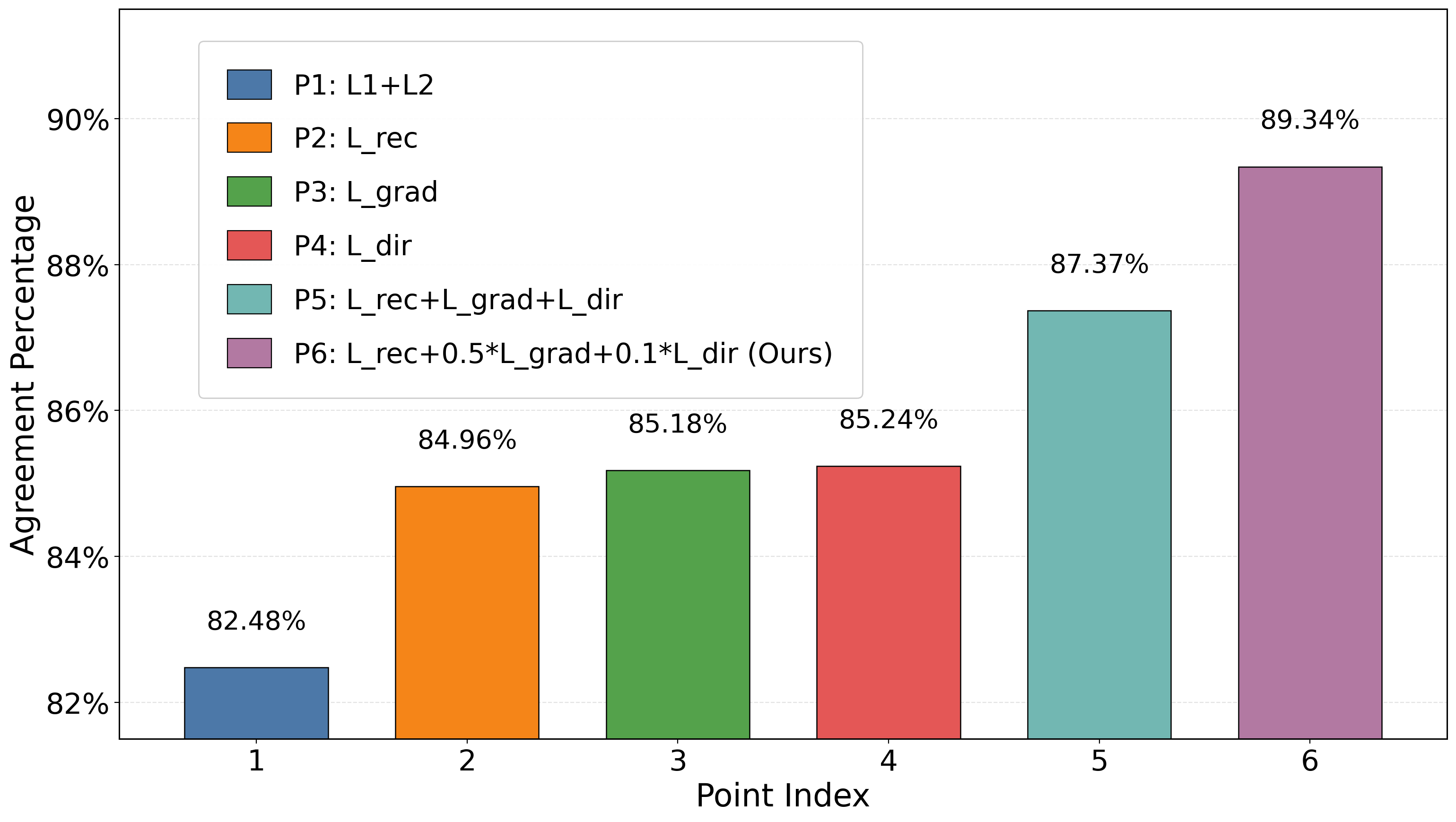}
}{%
\fbox{\parbox[c][0.16\textheight][c]{0.95\columnwidth}{\centering Placeholder for event-loss visualization.}}
}
\vspace{-3mm}
\caption{Visualization for event-domain supervision. The figure reports event-consistency statistics between the predicted background events and ground-truth events on a unified set of samples, where a pixel is counted as consistent if both event values are non-zero or both are zero under a near-zero threshold.}\vspace{-2mm}
\label{fig:ablation_event_loss_visualization}
\end{figure}

\begin{table}[!htb]
\centering
\caption{Ablation study on the fusion design. BF, RABF, RF, and EGRF denote background fusion, rain-aware background fusion, rain fusion, and event-guided rain fusion, respectively.}
\label{tab:ablation_fusion}
\setlength{\tabcolsep}{7pt}
\renewcommand{\arraystretch}{1.1}
\footnotesize
\begin{tabular}{cccccc}
\toprule
BF & RABF & RF & EGRF & PSNR$\uparrow$ & SSIM$\uparrow$ \\
\midrule
\cmark & \xmark & \xmark & \xmark & 38.12 & 0.9689 \\
\xmark & \cmark & \xmark & \xmark & 38.94 & 0.9728 \\
\xmark & \cmark & \cmark & \xmark & 39.61 & 0.9764 \\
\xmark & \cmark & \xmark & \cmark & 40.40 & 0.9789 \\
\bottomrule
\end{tabular}
\end{table}
Fig.~\ref{fig:ablation_event_loss_visualization} reports the consistency between predicted background events and ground-truth events. A pixel is regarded as consistent if both the prediction and ground truth are zero or both are non-zero under a near-zero threshold. Compared with the alternatives, the proposed loss yields higher agreement with the ground-truth event distribution. This indicates that the weighted event-domain supervision better preserves sparse event structures and reduces unreliable event responses.
% Fig.~\ref{fig:ablation_event_loss_visualization} visualizes the consistency between the predicted background events and ground-truth events under different losses on the same set of samples. For each pixel, we check whether the prediction and ground truth are both non-zero or both zero, where values below a near-zero threshold are treated as zero because of event sparsity. The consistency score is then computed as the ratio of matched pixels to all pixels. The visualization shows that our loss produces higher agreement with the ground-truth event distribution. This confirms that the proposed event-domain loss better preserves sparse event structures and provides more reliable event guidance for rain-background separation.

% \subsubsection{Fusion Design}

% We further evaluate the fusion strategy between RGB and event components. As shown in Table~\ref{tab:ablation_fusion}, using only background fusion achieves limited performance because it mainly compensates scene structures but does not explicitly enhance rain removal. Introducing rain-aware background fusion improves the result from 38.12 dB to 38.94 dB, showing that suppressing rain interference during background interaction is beneficial. When rain fusion is further added, the performance increases to 39.61 dB, which indicates that explicitly modeling rain components is necessary for effective deraining. Finally, replacing plain rain fusion with the proposed event-guided rain fusion achieves the best result of 40.40 dB and 0.9789 SSIM. These results demonstrate that background and rain components should be fused in a complementary manner: background fusion preserves scene content, while event-guided rain fusion strengthens the response to dynamic rain streaks. The consistent improvement across variants verifies the effectiveness of the proposed fusion design.
\subsubsection{Fusion Design}

We further study how RGB and event components should be fused. Table~\ref{tab:ablation_fusion} reports four variants. Using only background fusion achieves 38.12 dB, suggesting that background compensation alone is insufficient for removing dynamic rain streaks. Rain-aware background fusion improves the result to 38.94 dB by reducing rain interference during background interaction. Adding rain fusion further increases PSNR to 39.61 dB, which shows that explicitly modeling the rain component is necessary for deraining.

The proposed event-guided rain fusion obtains the best result, 40.40 dB and 0.9789 SSIM. Compared with plain rain fusion, it uses event-derived rain responses to strengthen the localization and suppression of dynamic rain streaks. These results suggest that background and rain components play complementary roles: background fusion mainly supports structure preservation, whereas event-guided rain fusion contributes to rain removal.

% \subsubsection{Number of Progressive Stages}

% Finally, we analyze the effect of the progressive stage number. As shown in Table~\ref{tab:ablation_stage_aggregation}, the one-stage model performs deraining in a single pass and therefore has limited ability to refine difficult rain residues. Increasing the depth to two stages improves the PSNR from 38.86 dB to 39.92 dB, and the three-stage model further reaches 40.40 dB and 0.9789 SSIM. Parameters and FLOPs increase nearly linearly with the number of stages, because each stage introduces a similar decoupling and interaction structure. Considering that the gain from two to three stages is already smaller than that from one to two stages, we do not further increase the stage number to four, which would bring extra cost with limited expected benefit.
\subsubsection{Number of Progressive Stages}

Finally, we evaluate the number of progressive stages. Table~\ref{tab:ablation_stage_aggregation} shows that a one-stage model obtains 38.86 dB. Increasing the number of stages to two improves PSNR to 39.92 dB, and the three-stage model further reaches 40.40 dB. The gain from one to two stages is 1.06 dB, while the gain from two to three stages is 0.48 dB. This indicates that progressive refinement is beneficial, but the marginal improvement decreases as the number of stages increases.

The parameters and FLOPs increase approximately linearly because each stage contains a similar decomposition and interaction structure. Considering the trade-off between accuracy and computation, we adopt three stages in the full model and do not further increase the stage number.

Fig.~\ref{fig:ablation_stage_visualization} shows stage-wise rain prediction results. The one-stage model captures only part of the rain pattern and includes some irrelevant background responses. The two-stage model predicts more complete streaks, while the three-stage model gives a cleaner rain map with fewer background activations. This supports the use of progressive refinement for gradually improving rain-background separation.
% The visualization in Fig.~\ref{fig:ablation_stage_visualization} gives a more intuitive explanation. From the one-stage prediction to the two-stage prediction, and further to the three-stage prediction, the estimated rain maps contain increasingly complete rain streaks. In particular, the three-stage model can correctly predict most rain streaks while suppressing irrelevant background responses. These results verify that progressive multi-stage modeling helps the network gradually enhance rain perception and leads to more reliable final restoration.

\begin{table}[!t]
\centering
\caption{Ablation study on the number of progressive stages.}
\label{tab:ablation_stage_aggregation}
\setlength{\tabcolsep}{7pt}
\renewcommand{\arraystretch}{1.1}
\footnotesize
\begin{tabular}{ccccc}
\toprule
Variant & PSNR$\uparrow$ & SSIM$\uparrow$ & Param & FLOPs \\
\midrule
One stage & 38.86 & 0.9741 & 4.12M & 349.12G \\
Two stages & 39.92 & 0.9776 & 7.16M & 607.98G \\
Three stages (Ours) & 40.40 & 0.9789 & 10.17M & 865.84G \\
\bottomrule
\end{tabular}
\end{table}

\begin{figure*}[!tbp]
\centering
\IfFileExists{ablation_stage_visualization.png}{%
\includegraphics[width=\textwidth]{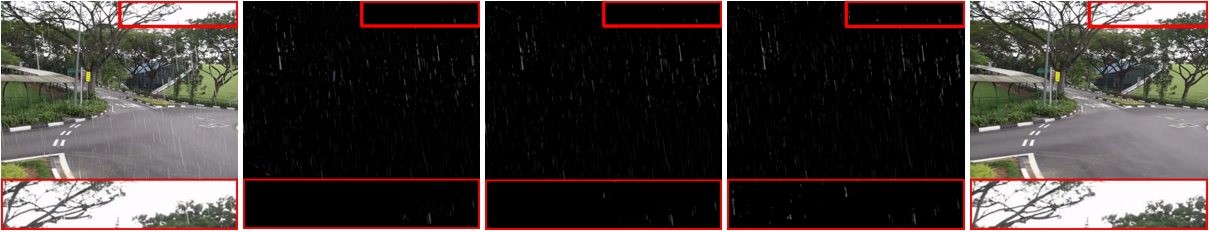}
}{%
\fbox{\parbox[c][0.16\textheight][c]{0.95\textwidth}{\centering Placeholder for stage-wise rain prediction visualization.}}
}
\vspace{0.5mm}
\setlength{\tabcolsep}{0pt}
\newcommand{\stagefiglabel}[2]{\makebox[0.2\textwidth][c]{\scriptsize\hspace{#1}#2}}
\noindent
\stagefiglabel{-8pt}{(a)}%
\stagefiglabel{-8pt}{(b)}%
\stagefiglabel{-8pt}{(c)}%
\stagefiglabel{-8pt}{(d)}%
\stagefiglabel{-8pt}{(e)}
\caption{Visualization for the progressive-stage ablation. (a) Rainy input, (b) rain map predicted by the one-stage model, (c) rain map predicted by the two-stage model, (d) rain map predicted by the three-stage model, and (e) ground truth.}\vspace{-4mm}
\label{fig:ablation_stage_visualization}
\end{figure*}

\begin{figure}[!htbp]
\centering
\includegraphics[width=\columnwidth]{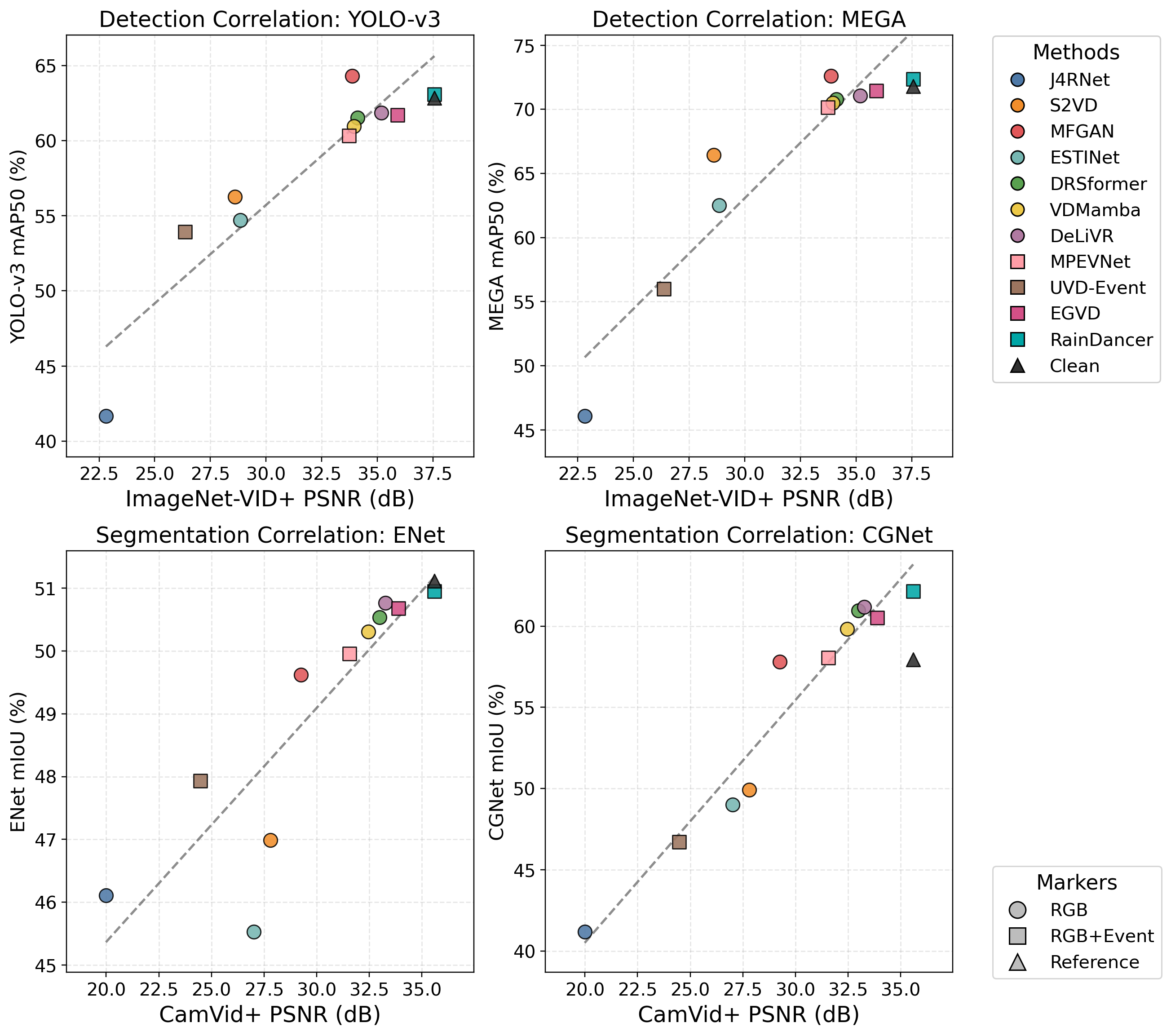}
\caption{Correlation between restoration quality and downstream perception on RainVID\&SS. The four subfigures show PSNR vs. YOLO-v3 mAP$_{50}$, PSNR vs. MEGA mAP$_{50}$, PSNR vs. ENet mIoU, and PSNR vs. CGNet mIoU, respectively.}
\label{fig:downstream_correlations}
\end{figure}

\subsection{Downstream Tasks}

We further evaluate whether deraining improves downstream perception on RainVID\&SS~\cite{ref2}, which contains ImageNet-VID+ for object detection and CamVid+ for semantic segmentation. For fair comparison, all deraining models are retrained on RainVID\&SS using the same experimental setting. We adopt YOLO-v3~\cite{ref78} and MEGA~\cite{ref79} as object detectors, and ENet~\cite{ref82} and CGNet~\cite{ref83} as semantic segmentation models.

Table~\ref{tab:downstream_results} reports both restoration and task performance. On ImageNet-VID+, our RainDancer method achieves the best PSNR/SSIM and the second-best detection accuracy for both YOLO-v3 and MEGA. MFGAN obtains slightly higher detection mAP, although its restoration metrics are lower than ours. This difference indicates that detection performance is not determined solely by full-reference restoration quality. Nevertheless, our RainDancer method improves the rainy input by a large margin and produces detection results close to or slightly higher than those obtained on clean frames under the tested detectors.

On CamVid+, our RainDancer method achieves the highest PSNR/SSIM and the best mIoU for both ENet and CGNet. The improvement is more evident for CGNet, where our RainDancer method reaches 62.14\% mIoU. These results suggest that component-aligned RGB-Event deraining helps preserve semantic boundaries and local structures that are useful for segmentation.

\begin{table*}[!tbp]
\centering
\caption{Object detection and semantic segmentation results on RainVID\&SS. PSNR and SSIM denote the full-reference restoration quality on the corresponding benchmark. $\mathrm{mAP}_{50}^{1}$ and $\mathrm{mAP}_{50}^{2}$ denote the object detection results obtained by YOLO-v3~\cite{ref78} and MEGA~\cite{ref79} on ImageNet-VID+, respectively. $\mathrm{mIoU}^{1}$ and $\mathrm{mIoU}^{2}$ denote the semantic segmentation results obtained by ENet~\cite{ref82} and CGNet~\cite{ref83} on CamVid+, respectively.\textit{Modality}: RGB denotes frame-only input, and RGB+Event denotes joint frame-event input.
The best and second-best results among deraining methods are highlighted in \textbf{bold} and \underline{underline}, respectively.}
\label{tab:downstream_results}
\setlength{\tabcolsep}{0pt}
\renewcommand{\arraystretch}{1.12}
\footnotesize

\resizebox{\textwidth}{!}{%
\begin{tabular}{@{}
c
@{\hspace{18pt}}
c
@{\hspace{18pt}}
c
@{\hspace{18pt}}
c
@{\hspace{18pt}}
c
@{\hspace{18pt}}
c
@{\hspace{18pt}}
c
@{\hspace{18pt}}
c
@{\hspace{18pt}}
c
@{\hspace{18pt}}
c
@{}}
\toprule
\multirow[c]{2}{*}{Method}
& \multirow[c]{2}{*}{Modality}
& \multicolumn{4}{c}{Object Detection (ImageNet-VID+)}
& \multicolumn{4}{c}{Semantic Segmentation (CamVid+)} \\
\cmidrule(l{0pt}r{18pt}){3-6} \cmidrule(l{0pt}r{0pt}){7-10}
&
& PSNR$\uparrow$
& SSIM$\uparrow$
& $\mathrm{mAP}_{50}^{1}\uparrow$
& $\mathrm{mAP}_{50}^{2}\uparrow$
& PSNR$\uparrow$
& SSIM$\uparrow$
& $\mathrm{mIoU}^{1}\uparrow$
& $\mathrm{mIoU}^{2}\uparrow$ \\
\midrule

Rainy (no processing)            & RGB       & 15.77 & 0.4027 & 21.23\% & 33.69\% & 18.64 & 0.4580 & 32.24\% & 16.80\% \\
(TPAMI 2020) J4RNet~\cite{ref8}  & RGB       & 22.81 & 0.7599 & 41.67\% & 46.08\% & 19.99 & 0.7562 & 46.11\% & 41.19\% \\
(CVPR 2021) S2VD~\cite{ref37}    & RGB       & 28.61 & 0.9007 & 56.27\% & 66.45\% & 27.81 & 0.8751 & 46.99\% & 49.93\% \\
(TPAMI 2022) MFGAN~\cite{ref76}  & RGB       & 33.88 & 0.9448 & \textbf{64.33\%} & \textbf{72.62\%} & 29.25 & 0.9091 & 49.62\% & 57.80\% \\
(TPAMI 2023) ESTINet~\cite{ref74} & RGB      & 28.85 & 0.7885 & 54.70\% & 62.52\% & 27.01 & 0.7863 & 45.53\% & 49.02\% \\
(CVPR 2023) DRSformer~\cite{ref72} & RGB     & 34.12 & 0.9392 & 61.54\% & 70.80\% & 32.98 & 0.9413 & 50.54\% & 60.97\% \\
(CVPR 2025) VDMamba~\cite{ref41} & RGB       & 33.96 & 0.9406 & 60.96\% & 70.52\% & 32.46 & 0.9368 & 50.31\% & 59.82\% \\
(ICLR 2026) DeLiVR~\cite{ref75}  & RGB       & 35.18 & 0.9456 & 61.88\% & 71.08\% & 33.26 & 0.9448 & \underline{50.77\%} & \underline{61.18\%} \\

\midrule

(TIP 2023) MPEVNet~\cite{ref2}   & RGB+Event & 33.74 & 0.9414 & 60.34\% & 70.14\% & 31.55 & 0.9252 & 49.96\% & 58.07\% \\
(ICCV 2023) UVD-Event~\cite{ref13} & RGB+Event & 26.36 & 0.8274 & 53.94\% & 56.01\% & 24.47 & 0.7993 & 47.93\% & 46.71\% \\
(TNNLS 2025) EGVD~\cite{ref12}   & RGB+Event & \underline{35.92} & \underline{0.9502} & 61.72\% & 71.46\% & \underline{33.88} & \underline{0.9481} & 50.68\% & 60.53\% \\
RainDancer (Ours)                      & RGB+Event & \textbf{37.57} & \textbf{0.9751} & \underline{63.08\%} & \underline{72.37\%} & \textbf{35.59} & \textbf{0.9597} & \textbf{50.95\%} & \textbf{62.14\%} \\

\midrule

Clean                    & RGB & Inf & 1.0000 & 62.84\% & 71.79\% & Inf & 1.0000 & 51.12\% & 57.96\% \\
\bottomrule
\end{tabular}%
}

\vspace{1mm}
\begin{minipage}{\textwidth}
\footnotesize
\end{minipage}
\end{table*}

% Fig.~\ref{fig:downstream_correlations} further shows that our method does not merely optimize PSNR. It achieves the highest PSNR while also obtaining strong detection and segmentation performance in all four settings, and even surpasses the clean-reference results except under ENet. This indicates that the proposed component-aligned RGB-Event interaction preserves task-relevant structures instead of over-smoothing them. The plots also show that higher PSNR does not always imply better perception: for example, S2VD and ESTINet obtain relatively high PSNR on CamVid+, but their ENet mIoU remains limited. This suggests that preserving semantic boundaries and stable local structures is crucial for downstream tasks.

Fig.~\ref{fig:downstream_correlations} further analyzes the relation between restoration quality and downstream accuracy. Our RainDancer method achieves the highest PSNR and consistently strong perception results, but the plots also show that higher PSNR does not always lead to higher mAP or mIoU. For example, some methods with competitive PSNR still obtain limited segmentation accuracy. This observation suggests that downstream perception depends not only on pixel-level fidelity but also on the preservation of semantic boundaries and task-relevant local structures. The proposed component-aligned interaction is beneficial in this respect because it reduces rain interference while maintaining background structures.

\begin{figure*}[!tbp]
\centering
\includegraphics[width=\textwidth]{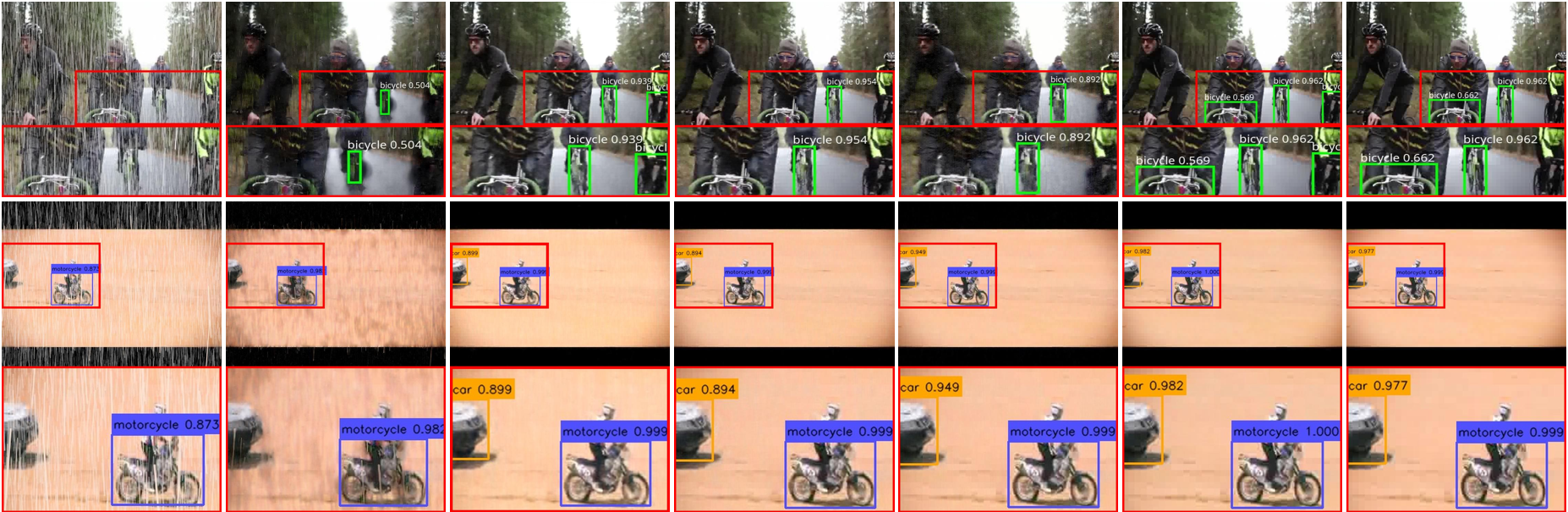}
\vspace{0.5mm}
\setlength{\tabcolsep}{0pt}
\newcommand{\detfiglabel}[2]{\makebox[0.1428\textwidth][c]{\scriptsize\hspace{#1}#2}}
\noindent
\detfiglabel{-8pt}{Rainy}%
\detfiglabel{-8pt}{J4RNet~\cite{ref8}}%
\detfiglabel{-8pt}{S2VD~\cite{ref37}}%
\detfiglabel{-10pt}{MPEVNet~\cite{ref2}}%
\detfiglabel{-8pt}{VDMamba~\cite{ref41}}%
\detfiglabel{-8pt}{RainDancer (Ours)}%
\detfiglabel{-8pt}{Clean}
\caption{Qualitative object detection results on ImageNet-VID+. The first and second rows show the detection results obtained by YOLO-v3 and MEGA, respectively.}\vspace{-4mm}
\label{fig:object_detection}
\end{figure*}

% Fig.~\ref{fig:object_detection} shows the qualitative object detection results. Under YOLO-v3, our restored frame enables the detector to identify the marginal bicycle that is missed in the clean image, indicating that deraining can enhance task-relevant local structures rather than merely improving visual appearance. Under MEGA, most methods can detect the motorcycle and car, but our method produces the highest confidence scores, showing more reliable detection after deraining.

Fig.~\ref{fig:object_detection} shows qualitative object detection results. In the YOLO-v3 example, the restored frame from our RainDancer method enables the detector to identify a marginal bicycle that is missed in several competing results. In the MEGA example, most methods detect the main objects, but our result yields higher confidence scores. These results indicate that deraining can improve object-level perception when the restored image preserves discriminative contours and local structures.

\begin{figure*}[!tbp]
\centering
\includegraphics[width=\textwidth]{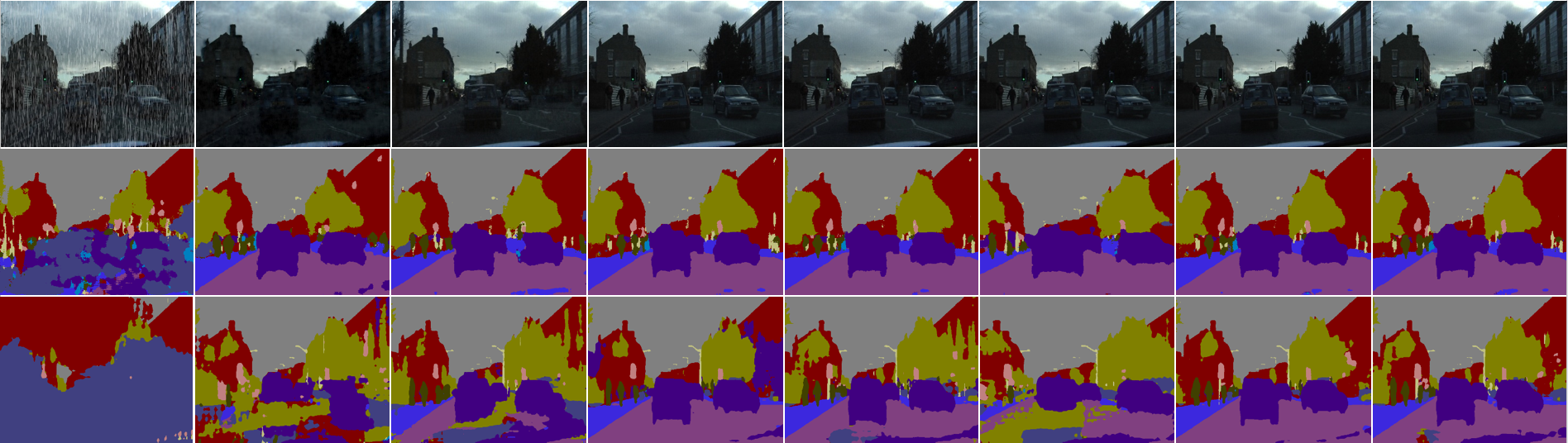}
\vspace{0.5mm}
\setlength{\tabcolsep}{0pt}
\newcommand{\segfiglabel}[2]{\makebox[0.125\textwidth][c]{\scriptsize\hspace{#1}#2}}
\noindent
\segfiglabel{-8pt}{Rainy}%
\segfiglabel{-8pt}{J4RNet~\cite{ref8}}%
\segfiglabel{-8pt}{S2VD~\cite{ref37}}%
\segfiglabel{-8pt}{EGVD~\cite{ref12}}%
\segfiglabel{-8pt}{MPEVNet~\cite{ref2}}%
\segfiglabel{-8pt}{DeLiVR~\cite{ref75}}%
\segfiglabel{-8pt}{RainDancer (Ours)}%
\segfiglabel{-8pt}{Clean}
\caption{Qualitative semantic segmentation results on CamVid+. The first row shows the derained images, while the second and third rows show the segmentation results obtained by ENet and CGNet, respectively.}\vspace{-4mm}
\label{fig:semantic_segmentation}
\end{figure*}

Fig.~\ref{fig:semantic_segmentation} presents semantic segmentation results on CamVid+. Compared with the competing methods, our restored images lead to more coherent semantic regions and clearer object boundaries for both ENet and CGNet. Several baselines either leave rain-induced disturbances or over-smooth local structures, which causes fragmented or unstable segmentation predictions. The results support the quantitative findings in Table~\ref{tab:downstream_results}: our RainDancer method improves restoration quality while better retaining structures that are useful for high-level visual understanding.
% Fig.~\ref{fig:semantic_segmentation} presents the qualitative semantic segmentation results. For both ENet and CGNet, the segmentation produced from our derained result is the closest to the clean-reference segmentation, with more coherent semantic regions and clearer object boundaries. In contrast, the competing methods still show obvious segmentation errors and unstable region predictions, indicating that their restorations do not provide sufficiently reliable structures for downstream understanding. This gap is particularly evident under CGNet. These observations further demonstrate that the proposed method better preserves task-relevant semantic structures for high-level vision tasks.

\section{Conclusion}
This paper presents a progressive RGB--Event video deraining framework following a \emph{decompose-before-interact} principle. Instead of directly fusing heterogeneous RGB and event features, the proposed method first decomposes each modality into rain and background components and then conducts component-level interaction between semantically aligned representations. A rain-oriented SNN event branch is developed to capture sparse and bursty rain dynamics, while rain-aware background fusion and event-guided rain fusion are introduced to reduce cross-modal interference and improve rain removal. Moreover, an event-domain supervision strategy is designed to constrain sparse reconstruction, structural consistency, and gradient orientation of event representations.
Experiments on synthetic and real RGB--Event video deraining datasets show that the proposed method outperforms existing RGB-based and event-guided approaches in quantitative metrics, visual quality, and downstream perception robustness. The results demonstrate that explicit rain-background decomposition and component-aligned RGB--Event interaction provide an effective solution for reducing rain-background entanglement in dynamic rainy videos.
%In this paper, we presented a progressive RGB-Event video deraining framework that follows a \emph{decompose-before-interact} design principle. Instead of directly mixing heterogeneous RGB and event features, the proposed method first decouples background and rain representations in each modality and then performs rain-aware cross-modal interaction between semantically aligned components. A rain-oriented SNN event disentanglement module was introduced to exploit sparse, bursty, and directionally coherent event rain dynamics, while conservative background and rain fusion modules enable reliable event-guided compensation without overwhelming the RGB restoration stream. In addition, structured event supervision further improves the physical faithfulness of event-background estimation. Extensive experiments on synthetic and real event-guided video deraining datasets demonstrate the effectiveness of the proposed framework in terms of restoration quality, model efficiency, and downstream perception. These results indicate that preserving modality-specific advantages before cross-modal fusion is a promising direction for robust video restoration under adverse weather.

% \section*{Acknowledgments}
% Acknowledgments are omitted in this version.

\bibliographystyle{IEEEtran}
\bibliography{refs}

\end{document}